\RequirePackage{fix-cm}

\documentclass[twocolumn]{svjour3}          % twocolumn
\smartqed  % flush right qed marks, e.g. at end of proof
\usepackage{graphicx}
\usepackage{cite}
\usepackage{times}
\usepackage{amsmath}
\usepackage{amssymb}
\usepackage{multirow}

\newcommand{\threefigures}[4]{
            \centerline{{\includegraphics[width=#4]{#1}}~~{\includegraphics[width=#4]{#2}}~~{\includegraphics[width=#4]{#3}}}}

\newcommand{\fourfigurescaption}[9]{
            \centerline{{\includegraphics[width=#5]{#1}}~~{\includegraphics[width=#5]{#2}}~~{\includegraphics[width=#5]{#3}}~~{\includegraphics[width=#5]{#4}}}
            \makebox[#5][c]{#6}\makebox[#5][c]{#7}\makebox[#5][c]{#8}\makebox[#5][c]{#9}}
            
\newcommand{\fivefigures}[6]{
            \centerline{{\includegraphics[width=#6]{#1}}~~{\includegraphics[width=#6]{#2}}~~{\includegraphics[width=#6]{#3}}~~{\includegraphics[width=#6]{#4}}~~{\includegraphics[width=#6]{#5}}}}

\newcommand{\fivefigurescaption}[6]{
            \centerline{{\includegraphics[width=#6]{#1}}~~{\includegraphics[width=#6]{#2}}~~{\includegraphics[width=#6]{#3}}~~{\includegraphics[width=#6]{#4}}~~{\includegraphics[width=#6]{#5}}}\makebox[#6][c]{Ground truth~~~~~~~~}\makebox[#6][c]{MP-Net-F~\cite{tokmakov2016learning}~~}\makebox[#6][c]{~~FSG~\cite{jain2017fusionseg}}\makebox[#6][c]{~~~~ARP~\cite{kohprimary}}\makebox[#6][c]{~~~~~~~~~Ours}}

\newcommand{\threefigurescaption}[7]{
            \centerline{{\includegraphics[width=#4]{#1}}~~{\includegraphics[width=#4]{#2}}~~{\includegraphics[width=#4]{#3}}}
		     \makebox[#4][c]{#5}\makebox[#4][c]{#6}\makebox[#4][c]{#7}}

\newcommand{\comment}[1]{}

\usepackage[pagebackref=false,breaklinks=true,letterpaper=true,colorlinks,bookmarks=false]{hyperref}

\begin{document}
\graphicspath{{figs/}}

\title{Learning to Segment Moving Objects}

\author{Pavel Tokmakov \and Cordelia Schmid \and Karteek Alahari}

\institute{The authors are at Inria.\\ \\
Univ.\ Grenoble Alpes, Inria, CNRS, Grenoble INP, LJK, 38000 Grenoble, France.\\
\email{firstname.lastname@inria.fr}
}

\date{Received: date / Accepted: date}

\maketitle

\begin{abstract}
We study the problem of segmenting moving objects in unconstrained videos.
Given a video, the task is to segment all the objects that exhibit independent motion in at least one frame. We formulate this as a learning problem and design our framework with three cues: (i)~independent object motion between a pair of frames, which complements object recognition, (ii)~object appearance, which helps to correct errors in motion estimation, and (iii)~temporal consistency, which imposes additional constraints on the segmentation. The framework is a two-stream neural network with an explicit memory module. The two streams encode appearance and motion cues in a video sequence respectively, while the memory module captures the evolution of objects over time, exploiting the temporal consistency. The motion stream is a convolutional neural network trained on synthetic videos to segment independently moving objects in the optical flow field. The module to build a ``visual memory'' in video, i.e., a joint representation of all the video frames, is realized with a convolutional recurrent unit learned from a small number of training video sequences.

For every pixel in a frame of a test video, our approach assigns an object or background label based on the learned spatio-temporal features as well as the ``visual memory'' specific to the video. We evaluate our method extensively on three benchmarks, DAVIS, Freiburg-Berkeley motion segmentation dataset and SegTrack. In addition, we provide an extensive ablation study to investigate both the choice of the training data and the influence of each component in the proposed framework.
\end{abstract}

\begin{figure*}
\begin{center}
\includegraphics[width=1.0\textwidth]{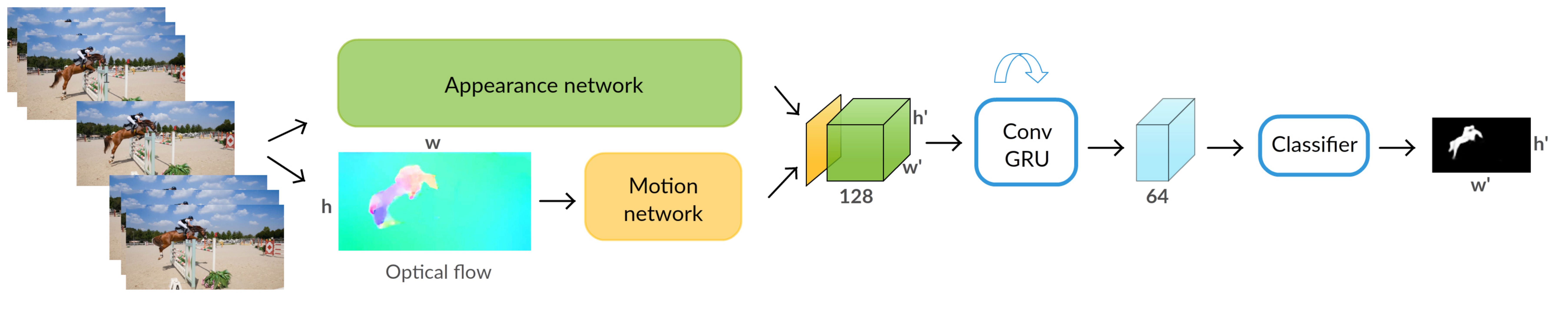}
\end{center}
\vspace{-0.3cm}
\caption{Overview of our segmentation approach. Each video frame is processed
by the appearance (green) and the motion (yellow) networks to produce an
intermediate two-stream representation. The ConvGRU module combines this with
the learned visual memory to compute the final segmentation result. The width
(w') and height (h') of the feature map and the output are $\text{w}/8$ and
$\text{h}/8$ respectively.}
\label{fig:model}
\end{figure*}

\section{Introduction}
\label{intro}
Video object segmentation is the task of extracting spatio-temporal regions
that correspond to object(s) moving in at least one frame in the video
sequence. The top-performing methods for this
problem~\cite{papazoglou2013fast,Faktor14} continue to rely on hand-crafted
features and do not leverage a learned video representation, despite the
impressive results achieved by convolutional neural networks (CNNs) for other
vision tasks, e.g., image segmentation~\cite{pinheiro2016learning}, object
detection~\cite{ren2015faster}. Very recently, there have been attempts to
build CNNs for video object
segmentation~\cite{jain2017fusionseg,Caelles17,Khoreva16}. Yet, they suffer
from various drawbacks. For example,~\cite{Caelles17,Khoreva16} rely on a
manually-segmented subset of frames (typically the first frame of the video
sequence) to guide the segmentation pipeline. The approach
of~\cite{jain2017fusionseg} does not require manual annotations, but remains
frame-based, failing to exploit temporal consistency in videos. Furthermore,
none of these methods has a mechanism to {\it memorize} relevant features of
objects in a scene. In this paper, we propose a novel framework to address
these issues.

We present a two-stream network with an explicit memory module for video object
segmentation (see Figure~\ref{fig:model}). The memory module is a convolutional
gated recurrent unit (GRU) that encodes the spatio-temporal evolution of
object(s) in the input video sequence. This spatio-temporal representation used
in the memory module is extracted from two streams---the appearance stream
which describes static features of objects in the video, and the temporal
stream which captures the independent object motion.

\begin{figure*}[t]
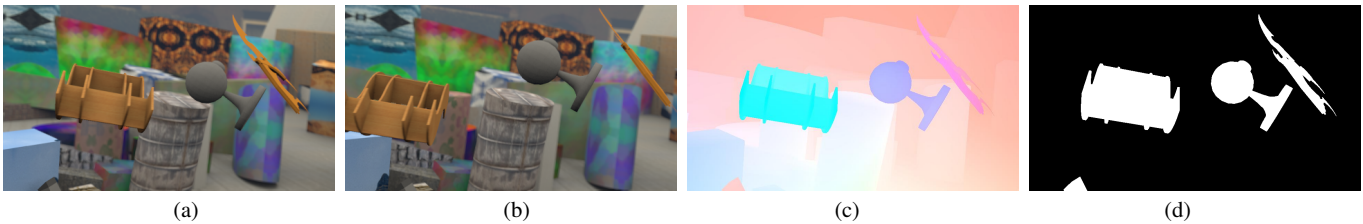

\begin{center}
\fourfigurescaption{2/exfrm1}{2/exfrm2}{2/exflow}{2/exgt}{0.25\textwidth}{(a)}{(b)}{(c)}{(d)}
\end{center}
\vspace{-0.3cm}\caption{(a,b) Two example frames from a sequence in the
FlyingThings3D dataset~\cite{Mayer16}. The camera is in motion in this scene,
along with four independently moving objects. (c) Ground-truth optical flow of
(a), which illustrates motion of both foreground objects and background with
respect to the next frame (b). (d) Ground-truth segmentation of moving objects
in this scene.}
\label{fig:intro}
\end{figure*}

The temporal stream separates independent object and camera motion with our
motion pattern network (MP-Net), a trainable model, which takes optical flow as
input and outputs a per-pixel score for moving objects. Inspired by fully
convolutional networks (FCNs)~\cite{long2015fully,Dosovitskiy15,Ronneberger15},
we propose a related encoder-decoder style architecture to accomplish this
two-label classification task. The network is trained from scratch with
synthetic data~\cite{Mayer16}.  Pixel-level ground-truth labels for training
are generated automatically (see Figure~\ref{fig:intro}(d)), and denote whether
each pixel has moved in the scene. The input to the network is flow fields,
such as the one shown in Figure~\ref{fig:intro}(c). More details of the
network, and the training procedure are provided in
Section~\ref{sec:trainmpnet}. With this training, our model learns to
distinguish motion patterns of objects and background.

The appearance stream is the DeepLab
network~\cite{chen2014semantic,CP2016Deeplab}, pretrained on the PASCAL VOC
segmentation dataset, and it operates on individual video frames.  With the
spatial and temporal CNN features, we train the convolutional GRU component of
the framework to learn a {\it visual memory} representation of object(s) in the
scene. Given a frame $t$ from the video sequence as input, the network extracts
its spatio-temporal features and: (i) computes the segmentation using the
memory representation aggregated from all frames previously seen in the video,
(ii) updates the memory unit with features from $t$. The segmentation is
improved further by processing the video in the forward and the backward
directions in the memory unit, with our {\it bidirectional convolutional GRU}.

The contributions of the paper are three-fold. First we demonstrate that
independent motion between a pair of frames can be learned, and emphasize the
utility of synthetic data for this task (see \S\ref{sec:mpnet}). Second, we present an approach for
moving object segmentation in unconstrained videos that does not require any
manually-annotated frames in the input video (see \S\ref{sec:fullmodel}).  Our
network architecture incorporates a memory unit to capture the evolution of
object(s) in the scene (see \S\ref{sec:recurrent}). To our knowledge, this is
the first recurrent network based approach for the video segmentation task. It
helps address challenging scenarios where the motion patterns of the object
change over time; for example, when an object in motion stops to move,
abruptly, and then moves again, with potentially a different motion pattern.
Finally, we present state-of-the-art results on the DAVIS~\cite{Perazzi16} and
Freiburg-Berkeley motion segmentation (FBMS)~\cite{ochs2014segmentation}
benchmark datasets, and competitive results on SegTrack-v2~\cite{li2013video}
(see \S\ref{sec:soa}). We also provide an extensive experimental analysis, with
ablation studies to investigate the influence of all the components of our
framework in Section~\ref{sec:abl}. 

Preliminary versions of this work have been published at
CVPR~\cite{tokmakov2016learning} and ICCV~\cite{tokmakov2017learning}. Here, we extend these previous publications by: (i) significantly improving
the performance of MP-Net with better optical flow estimation and finetuning
the network on real videos (see \S\ref{sec:realvidexp} and
\S\ref{sec:realvidtrain}), (ii) replacing the DeepLab
v1~\cite{chen2014semantic} appearance stream in our moving object segmentation
framework with the ResNet-based DeepLab v2~\cite{CP2016Deeplab} and showing
that this indeed improves the performance (see \S\ref{sec:abl}), (iii)
studying the effect of motion estimation quality on the overall segmentation
results (see \S\ref{sec:mos}), and (iv) providing an analysis of the learned
spatio-temporal representation (see \S\ref{sec:gru}).

Scoure code and trained models are  available
online at \url{http://thoth.inrialpes.fr/research/lvo}.

\section{Related work}
\label{sec:rel}
Our work is related to: motion and scene flow estimation, video object
segmentation, and recurrent neural networks. We will review the most relevant
work on these topics in the remainder of this section.

\paragraph{\bf Motion estimation.} Early attempts for estimating motion have
focused on geometry-based approaches, such as~\cite{Torr98}, where the
potential set of motions is identified with RANSAC. Recent methods have relied
on other cues to estimate moving object regions. For example, Papzouglou and
Ferrari~\cite{papazoglou2013fast} first extract motion boundaries by measuring
changes in the optical flow field, and use it to estimate moving regions.  They
also refine this initial estimate iteratively with appearance features.  This
approach produces interesting results, but is limited by its heuristic
initialization. We show that incorporating our learning-based motion estimation
into it improves the results significantly (see Table~\ref{tbl:bms}).

Narayana~et al.~\cite{Narayana13} use optical flow orientations in a
probabilistic model to assign per-pixel labels that are consistent with their
respective real-world motion. This approach assumes pure translational camera
motion, and is prone to errors when the object and camera motions are
consistent with each other. Bideau~et al.~\cite{Bideau16} presented an
alternative to this, where initial estimates of foreground and background
motion models are updated over time, with optical flow orientations of the new
frames. This initialization is also heuristic, and lacks a robust learning
framework.  While we also set out with the goal of finding objects in motion, our
solution to this problem is a novel learning-based method. Scene flow, i.e., 3D
motion field in a scene~\cite{Vedula05}, is another form of motion estimation,
but is computed with additional information, such as disparity values computed
from stereo images~\cite{Huguet07,Wedel11}, or estimated 3D scene
models~\cite{Vogel15}. None of these methods follows a CNN-based learning
approach, in contrast to our method.

In concurrent work, Jain~et al.~\cite{jain2017fusionseg} presented a deep
network to segment independent motion in the flow field. While their approach
is related to ours, they use frame pairs from real videos, in contrast to
synthetic data in our case. Consequently, their work relies on estimated
optical flow in training. Since obtaining accurate ground truth moving object
segmentation labels is prohibitively expensive for a large dataset, they rely
on an automatic, heuristic-based label estimation approach, which results in
noisy annotations. We explore the pros and cons of using this realistic but
noisy dataset for training our motion segmentation model in
Section~\ref{sec:realvidtrain}.

\paragraph{\bf Video object segmentation.} The task of segmenting objects in
video is to associate pixels belonging to a class spatio-temporally; in other
words, extract segments that respect object boundaries, as well as associate
object pixels temporally whenever they appear in the video. This can be
accomplished by propagating manual segment labels in one or more frames to the
rest of the video sequence~\cite{Badrinarayanan10}. This class of methods is
not applicable to our scenario, where no manual segmentation is available.

Our approach to solve the segmentation problem does not require any
manually-marked regions. Several methods in this paradigm generate an
over-segmentation of
videos~\cite{Brendel09,Grundmann10,Lezama11,XuC16,Khoreva15}. While this can be
a useful intermediate step for some recognition tasks in video, it has no
notion of objects. Indeed, most of the extracted segments in this case do not
directly correspond to objects, making it non-trivial to obtain video object
segmentation from this intermediate result. An alternative to this is
clustering pixels spatio-temporally based on motion features computed along
individual point
trajectories~\cite{brox2010object,fragkiadaki2012video,ochs2012higher}, which
produces more coherent regions. They, however, assume homogeneity of motion
over the entire object, which is invalid for non-rigid objects.

Another class of segmentation methods casts the problem as a
foreground-background classification
task~\cite{Faktor14,papazoglou2013fast,wang2015saliency,taylor2015causal,zhang2013video,lee2011key}.
Some of these first estimate a
region~\cite{papazoglou2013fast,wang2015saliency} or
regions~\cite{lee2011key,zhang2013video}, which potentially correspond(s) to
the foreground object, and then learn foreground/background appearance models.
The learned models are then integrated with other cues, e.g., saliency
maps~\cite{wang2015saliency}, pairwise
constraints~\cite{papazoglou2013fast,zhang2013video}, object shape
estimates~\cite{lee2011key}, to compute the final object segmentation.
Alternatives to this framework have used: (i)~long-range interactions between
distinct parts of the video to overcome noisy initializations in low-quality
videos~\cite{Faktor14}, and (ii)~occluder/occluded relations to obtain a
layered segmentation~\cite{taylor2015causal}. While our proposed method is
similar in spirit to this class of approaches, in terms of formulating
segmentation as a classification problem, we differ from previous work
significantly. We propose an integrated approach to learn appearance and motion
features, and update them with a memory module, in contrast to estimating an
initial region heuristically and then propagating it over time. Our robust
model outperforms all the top ones from this
class~\cite{papazoglou2013fast,wang2015saliency,lee2011key,taylor2015causal,Faktor14},
as shown in Section~\ref{sec:soa}.

Very recently, CNN-based methods for video object segmentation were
proposed~\cite{Caelles17,Khoreva16,jain2017fusionseg}. Starting with CNNs
pretrained for image segmentation, two of these
methods~\cite{Caelles17,Khoreva16} find objects in video by finetuning on the
first frame in the sequence. Note that this setup, referred to as
semi-supervised segmentation, is very different from the more challenging
unsupervised case we address in this paper, where no manually-annotated frames
are available for the test video. Furthermore, these two CNN architectures are
primarily developed for images, and do not model temporal information in video.
We, on the other hand, propose a recurrent network specifically for the video
segmentation task. Jain~et al.~\cite{jain2017fusionseg} augment their motion
segmentation network with an appearance model and learn the parameters of a
layer to combine the predictions of the two. Their model does not feature a
memory module, and also remains frame-based. Thus, it can not exploit the
temporal consistency in video. We outperform~\cite{jain2017fusionseg} on DAVIS
and FBMS.
\begin{figure*}[t]
\begin{center}
\includegraphics[width=1.0\textwidth]{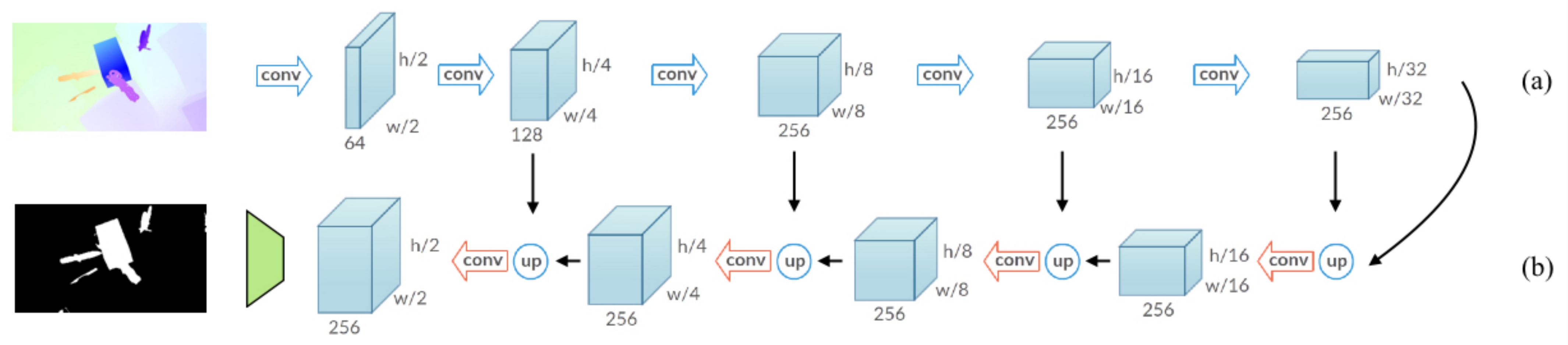}
\end{center}
\caption{Our motion pattern network: MP-Net. The blue arrows in the encoder
part (a) denote convolutional layers, together with ReLU and max-pooling
layers. The red arrows in the decoder part (b) are convolutional layers with
ReLU, `up' denotes $2\times2$ upsampling of the output of the previous unit.
The unit shown in green represents bilinear interpolation of the output of the
last decoder unit. }
\label{fig:mpnet}
\end{figure*}

\paragraph{\bf Recurrent neural networks (RNNs).}
RNN~\cite{hopfield1982neural,rumelhart86} is a popular model for tasks defined
on sequential data. Its main component is an internal state that allows to
accumulate information over time. The internal state in classical RNNs is
updated with a weighted combination of the input and the previous state, where
the weights are learned from training data for the task at hand. Long
short-term memory (LSTM)~\cite{hochreiter1997long} and GRU~\cite{Cho14}
architectures are improved variants of RNN, which partially mitigate the issue
of vanishing gradients~\cite{pascanu2013difficulty,Hochreiter98}. They
introduce gates with learnable parameters, to update the internal state
selectively, and can propagate gradients further through time.

Recurrent models, originally used for text and speech recognition,
e.g.,~\cite{graves2013speech,mikolov2010recurrent}, are becoming increasingly
popular for visual data. Initial work on vision tasks, such as image
captioning~\cite{donahue2015long}, future frame
prediction~\cite{srivastava2015unsupervised} and action
recognition~\cite{NgHVVMT15}, has represented the internal state of the
recurrent models as a 1D vector---without encoding any spatial information.
LSTM and GRU architectures have been extended to address this issue with the
introduction of ConvLSTM~\cite{xingjian2015convolutional,patraucean2015spatio,finn2016unsupervised}
and ConvGRU~\cite{ballas2015delving} respectively. In
these convolutional recurrent models the state and the gates are 3D tensors and
the weight vectors are replaced by 2D convolutions. These models have only
recently been applied to vision tasks, such as video frame
prediction~\cite{finn2016unsupervised,patraucean2015spatio,xingjian2015convolutional},
action recognition and video captioning~\cite{ballas2015delving}.

In this paper, we employ a visual memory module based on a convolutional GRU
(ConvGRU), and show that it is an effective way to encode the
spatio-temporal evolution of objects in video for segmentation. Further, to
fully benefit from all the frames in a video sequence, we apply the recurrent
model bidirectionally~\cite{graves2005framewise,graves2013hybrid}, i.e., apply
two identical model instances on the sequence in forward and backward
directions, and combine the predictions for each frame. This makes our memory
module a {\it bidirectional convolutional recurrent model}.

\section{Learning to segment moving objects in videos}
\label{sec:fullmodel}
We start by describing the overall architecture of our video object
segmentation framework. It takes video frames together with their estimated
optical flow as input, and outputs binary segmentations of moving objects, as
shown in Figure~\ref{fig:model}. We target the most general form of this task,
wherein objects are to be segmented in the entire video if they move in at
least one frame. The proposed model is comprised of three key components:
appearance and motion networks, and a visual memory module described below.

\paragraph{\bf Appearance network.}
The purpose of the appearance stream is to produce a high-level encoding of a
frame that will later aid the visual memory module in forming a representation
of the moving object. It takes a $\text{w} \times \text{h}$ RGB frame as input
and produces a $128 \times \text{w}/8 \times \text{h}/8$ feature representation
(shown in green in Figure~\ref{fig:model}), which encodes the semantic content
of the scene. As a  baseline for this stream we use the largeFOV, VGG16-based
version of the DeepLab network~\cite{chen2014semantic}. This network's
architecture is based on dilated convolutions~\cite{chen2014semantic}, which
preserve a relatively high spatial resolution of features, and also incorporate
context information in each pixel's representation. It is pretrained on a
semantic segmentation dataset~\cite{pascalvoc2012}, resulting in features that
can distinguish objects from background as well as from each other---a crucial
aspect for the video object segmentation task. We also experiment (in
\S\ref{sec:abl}) with upgrading the appearance stream to
DeepLab-v2~\cite{CP2016Deeplab}, a more recent version of the model, where the
VGG16 architecture is replaced with ResNet101, and the network is additionally
pretrained on the COCO semantic segmentation dataset~\cite{lin2014microsoft}.

\paragraph{\bf Motion network.}
For the temporal stream we employ a CNN pretrained for the motion segmentation
task. It is trained to estimate independently moving objects (i.e.,
irrespective of camera motion) based on optical flow computed from a pair of
frames as input; see Section~\ref{sec:mpnet} for details. This stream (shown in
yellow in Figure~\ref{fig:model}) produces a $\text{w}/4 \times \text{h}/4$
motion prediction output, where each value represents the likelihood of the
corresponding pixel being in motion. Its output is further downsampled by a
factor 2 (in w and h) to match the dimension of the appearance stream output.

The intuition behind using two streams is to benefit from their complementarity
for building a strong representation of objects that evolves over time. For
example, both appearance and motion networks are equally effective when an
object is moving in the scene, but as soon as it becomes stationary, the motion
network can not estimate the object, unlike the appearance network. We leverage
this complementary nature, as done by two-stream networks for other vision
tasks~\cite{simonyan2014two}. Note that our approach is not specific to the
particular networks described above, but is in fact a general framework for
video object segmentation. As shown is the Section~\ref{sec:abl}, its
components can easily replaced with other networks, providing scope for future
improvement.

\paragraph{\bf Memory module.}
The third component, i.e., a visual memory module, takes the concatenation of appearance and motion stream outputs as its input. It refines the initial
estimates from these two networks, and also memorizes the appearance and
location of objects in motion to segment them in frames where: (i) they are
static, or (ii) motion prediction fails. The output of this ConvGRU memory
module is a $64 \times \text{w}/8 \times \text{h}/8$ feature map obtained by
combining the two-stream input with the internal state of the memory module, as
described in detail in Section~\ref{sec:recurrent}. We further improve the
model by processing the video bidirectionally; see Section~\ref{sec:bidirec}.
The output from the ConvGRU module is processed by a $1 \times 1$ convolutional
layer and a softmax nonlinearity to produce the final pairwise segmentation
result.

\section{Motion pattern network}
\label{sec:mpnet}
Our MP-Net takes the optical flow field corresponding to two consecutive frames
of a video sequence as input, and produces per-pixel motion labels. We treat
each video as a sequence of frame pairs, and compute the labels independently
for each pair. As shown in Figure~\ref{fig:mpnet}, the network comprises
several ``encoding'' (convolutional and max-pooling) and ``decoding''
(upsampling and convolutional) layers. The motion labels are produced by the
last layer of the network, which are then rescaled to the original image
resolution (see \S\ref{sec:net}). We train the network on synthetic
data---a scenario where ground-truth motion labels can be acquired easily (see
\S\ref{sec:trainmpnet}). We also experiment with finetuning our MP-Net on real videos (see \S\ref{sec:realvidtrain}). For a detailed discussion of  motion patterns our approach detects refer to \S\ref{sec:realvid}.

\subsection{Network architecture}
\label{sec:net}
Our encoder-decoder style network is motivated by the goal of segmenting
diverse motion patterns in flow fields, which requires a large receptive field
as well as an output at the original image resolution. A large receptive field
is critical to incorporate context into the model. For example, when the
spatial region of support (for performing convolution) provided by a small
receptive field falls entirely within an object with non-zero flow values, it
is impossible to determine whether it is due to object or camera motion. On the
other hand, a larger receptive field will include regions corresponding to the
object as well as background, providing sufficient context to determine what is
moving in the scene. The second requirement of output generated at the original
image resolution is to capture fine details of objects, e.g., when only a part
of the object is moving. Our network satisfies these two requirements with: (i)
the encoder part learning features with receptive fields of increasing sizes,
and (ii) the decoder part upsampling the intermediate layer outputs to finally
predict labels at the full resolution. This approach is inspired by recent advances in semantic segmentation, where similar requirements are encountered~\cite{Ronneberger15}.

Figure~\ref{fig:mpnet} illustrates our network architecture. Optical flow
field input is processed by the encoding part of the network (denoted by (a) in
the figure) to generate a coarse representation that is a $32\times32$
downsampled version of the input. Each 3D block here represents a feature map
produced by a set of layers. In the encoding part, each feature map is a result
of applying convolutions, followed by a ReLU non-linearity layer, and then a
$2\times2$ max-pooling layer. The coarse representation learned by the final set
of operations in this part, i.e., the $32\times32$ downsampled version, is
gradually upsampled by the decoder part ((b) in the figure). In each decoder
step, we first upsample the output of the previous step by $2\times2$, and
concatenate it with the corresponding intermediate encoded representation,
before max-pooling (illustrated with black arrows pointing down in the figure).
This upscaled feature map is then processed with two convolutional layers,
followed by non-linearities, to produce input for the next (higher-resolution)
decoding step. The final decoder step produces a motion label map at half the
original resolution. We perform a bilinear interpolation on this result to
estimate labels at the original resolution.

\subsection{Training with synthetic data}
\label{sec:trainmpnet}
We need a large number of fully-labelled examples to train a convolutional
network such as the one we propose. In our case, this data corresponds to
videos of several types of objects, captured under different conditions (e.g.,
moving or still camera), with their respective moving object annotations. No
large dataset of real-world scenes satisfying these requirements is currently
available, predominantly due to the cost of generating ground-truth annotations
and flow for every frame. We adopt the popular approach of using synthetic
datasets, followed in other work~\cite{Dosovitskiy15,Mayer16}.
Specifically, we use the FlyingThings3D dataset~\cite{Mayer16} containing 2250
video sequences of several objects in motion, with ground-truth optical flow.
We augment this dataset with ground-truth moving object labels, which are
accurately estimated using the disparity values and camera parameters available
in the dataset, as outlined in Section~\ref{sec:dataset}. See
Figure~\ref{fig:intro}(d) for an illustration.

We train the network with mini-batch SGD under several settings. The one
trained with ground-truth optical flow as input shows the best performance.
This is analyzed in detail in Section~\ref{sec:mod}. Note that, while we use
ground-truth flow for training and evaluating the network on synthetic
datasets, all our results on real-world test data use only the estimated
optical flow. After convergence of the training procedure, we obtain a learned
model for motion patterns.

Our approach capitalizes on the recent success of CNNs for pixel-level
labeling tasks, such as semantic image segmentation, which learn feature
representations at multiple scales in the RGB space. The key to their top
performance is the ability to capture local patterns in images. Various types
of object and camera motions also produce consistent local patterns in the flow
field, which our model is able to learn to recognize. This gives us a clear
advantage over other pixel-level motion estimation
techniques~\cite{Bideau16,Narayana13} that can not detect local patterns.
Motion boundary based heuristics used in~\cite{papazoglou2013fast} can be seen
as one particular type of pattern, representing independent object motion. Our
model is able to learn many such patterns, which greatly improves the quality
and robustness of motion estimation.

\subsection{Detecting motion patterns}
\label{sec:realvid}
\begin{figure}[t]
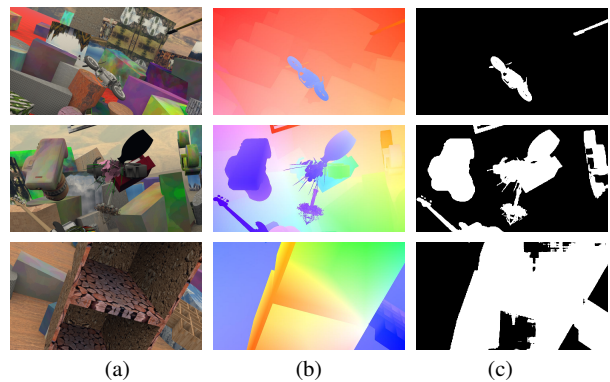

\begin{center}
\threefigures{6/exfrm1}{6/exflow}{6/exresult}{0.3\columnwidth} \vspace{0.1cm}
\threefigures{3/exfrm1}{3/exflow}{3/exresult}{0.3\columnwidth} \vspace{0.1cm}
\threefigurescaption{8/exfrm1}{8/exflow}{8/exresult}{0.3\columnwidth}{(a)}{(b)}{(c)}
\end{center}
\vspace{-0.3cm}\caption{Each row shows: (a) example frame from a sequence in
FlyingThings3D, (b) ground-truth optical flow of (a), which illustrates motion
of both foreground objects and background, with respect to the next frame, and
(c) our estimate of moving objects in this scene with ground-truth optical flow
as input.}
\label{fig:qualff}
\end{figure}

We apply our trained model on synthetic (FlyingThings3D) as well as real-world
(DAVIS, FBMS, SegTrack-v2) test data. Figure~\ref{fig:qualff} shows sample
predictions of our model on the FlyingThings3D test set with ground-truth
optical flow as input. Examples in the first two rows show that our model
accurately identifies fine details in objects: thin structures even when they
move subtlely, such as the neck of the guitar in the top-right corner in the
first row (see the subtle motion in the optical flow field (b)), fine
structures like leaves in the vase, and the guitar's headstock in the second
row.  Furthermore, our method successfully handles objects exhibiting highly
varying motions in the second example. The third row shows a limiting case,
where the receptive field of our network falls entirely within the interior of
a large object, as the moving object dominates. Traditional approaches, such as
RANSAC, do not work in this case either.

\begin{figure*}[t]
\begin{center}
\begin{tabular}{ccc}
\includegraphics[width=0.3\textwidth]{objectness/frame2} & 
\includegraphics[width=0.3\textwidth]{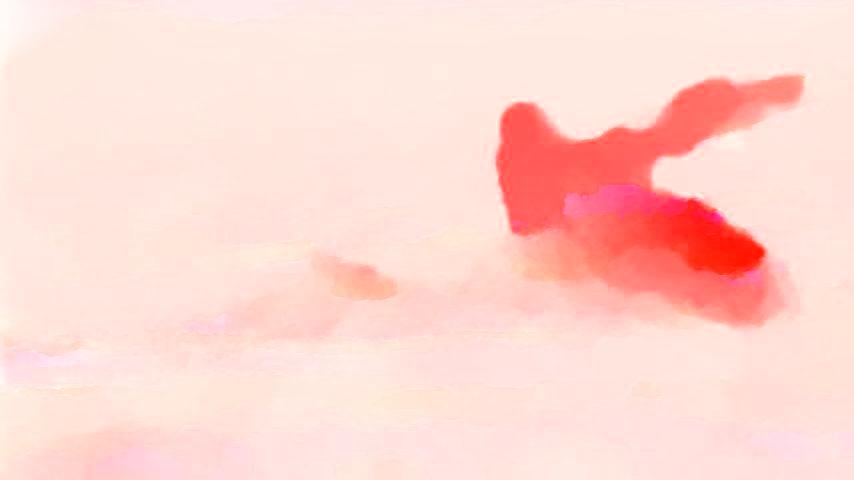} & 
\includegraphics[width=0.3\textwidth]{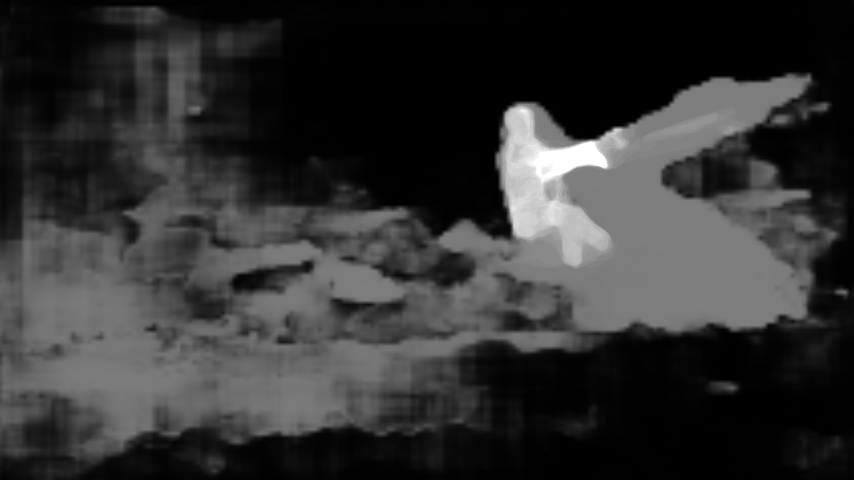} \\ 

\includegraphics[width=0.3\textwidth]{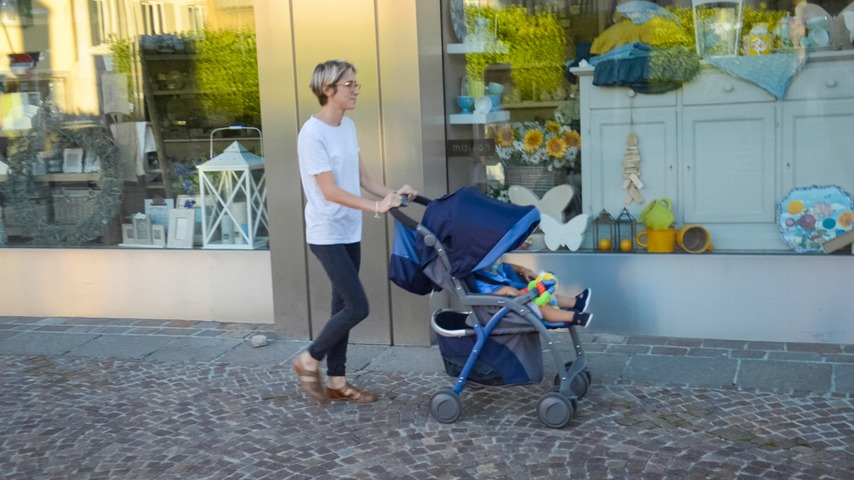} & 
\includegraphics[width=0.3\textwidth]{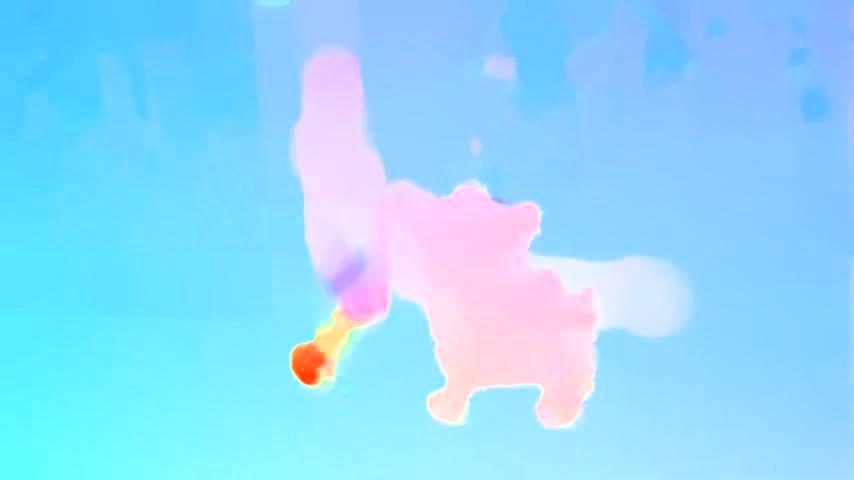} & 
\includegraphics[width=0.3\textwidth]{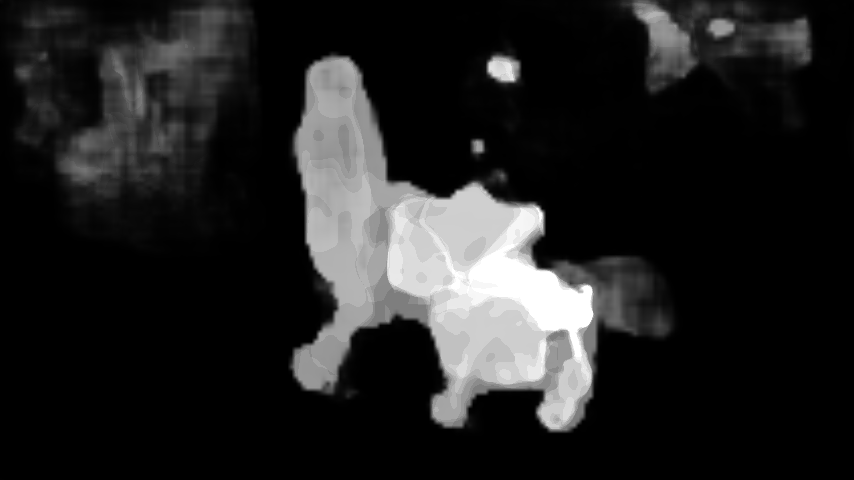}  \\ 
(a) & (b) & (c) 
\end{tabular}
\end{center}
\vspace{-0.3cm}\caption{Sample results on the DAVIS dataset for MP-Net. Each row shows: (a) video frame, (b) optical flow
estimated with LDOF~\cite{Brox11a}, (c) output of our MP-Net with LDOF
flow as input.}%\vspace{-0.5cm}}
\label{fig:obj}
\end{figure*}

In order to detect motion patterns in real-world videos, we first compute
optical flow with popular
methods~\cite{sundaram2010dense,Revaud15,ilg2016flownet}. With this
flow as input to the network, we estimate a motion label map, as shown in the
examples in Figure~\ref{fig:obj}(c). Although the prediction of our frame-pair
feedforward model is accurate in several regions in the frame ((c) in the
figure), we are faced with two challenges, which were not observed in the
synthetic training set. The first one is motion of
\textit{stuff}~\cite{Adelson01} in a scene, e.g., patterns on the water due to
the kiteboarder's motion (first row in the figure), which is irrelevant for
moving object segmentation. The second one is significant errors in optical
flow, e.g., in front of the pram ((b) in the bottom row in the figure).
Furthermore, this motion segmentation approach is purely frame-based, thus
unable to exploit temporal consistency in a video, and does not segment object
in frames where they stop moving. In our previous
work~\cite{tokmakov2016learning} we introduced post-processing steps to handle
some of these problems. In particular, we incorporated an objectness map
computed with object proposals~\cite{pinheiro2016learning} to suppress motion
corresponding to stuff, as well as false positives due to errors in flow
estimation. This post-processing allowed the method to achieve competitive
results, but it remained frame-level. The video object segmentation framework
presented in this paper addresses all these issues, as shown experimentally in
Section~\ref{sec:soa}.
\setlength{\tabcolsep}{2pt}
\setlength{\tabcolsep}{6pt}

\section{ConvGRU visual memory module}
\label{sec:recurrent}
The key component of the ConvGRU module is the state matrix $h$, which encodes
the visual memory. For frame $t$ in the video sequence, ConvGRU uses the
two-stream representation $x_t$ and the previous state $h_{t-1}$ to compute the
new state $h_t$. The dynamics of this computation are guided by an update gate
$z_t$, a forget gate $r_t$.
The states and the gates are 3D tensors, and can characterize spatio-temporal
patterns in the video, effectively memorizing which objects move, and where
they move to. These components are computed with convolutional operators and
nonlinearities as follows.
\begin{eqnarray}
	z_t &=& \sigma(x_t * w_{xz} + h_{t-1} * w_{hz} + b_{z}), \label{eqn:update} \\
	r_t &=& \sigma(x_t * w_{xr} + h_{t-1} * w_{hr} + b_{r}), \label{eqn:reset} \\
	\tilde{h}_t &=& \tanh(x_t * w_{x\tilde{h}} + r_t \odot h_{t-1} * w_{h\tilde{h}} + b_{\tilde{h}}), \label{eqn:candmem} \\
	h_t &=& (1 - z_t) \odot h_{t-1}  + z_t \odot \tilde{h}_t, \label{eqn:state}
\end{eqnarray}
where $\odot$ denotes element-wise multiplication, $*$ represents a
convolutional operation, $\sigma$ is the sigmoid function, $w$'s are learned
transformations, and $b$'s are bias terms.

The new state $h_t$ in (\ref{eqn:state}) is a weighted combination of the
previous state $h_{t-1}$ and the candidate memory $\tilde{h}_t$. The update
gate $z_t$ determines how much of this memory is incorporated into the new
state. If $z_t$ is close to zero, the memory represented by $\tilde{h}_t$ is
ignored. The reset gate $r_t$ controls the influence of the previous state
$h_{t-1}$ on the candidate memory $\tilde{h}_t$ in (\ref{eqn:candmem}), i.e.,
how much of the previous state is let through into the candidate memory. If
$r_t$ is close to zero, the unit forgets its previously computed state
$h_{t-1}$.
\begin{figure}[t]
\begin{center}
\includegraphics[width=0.85\columnwidth]{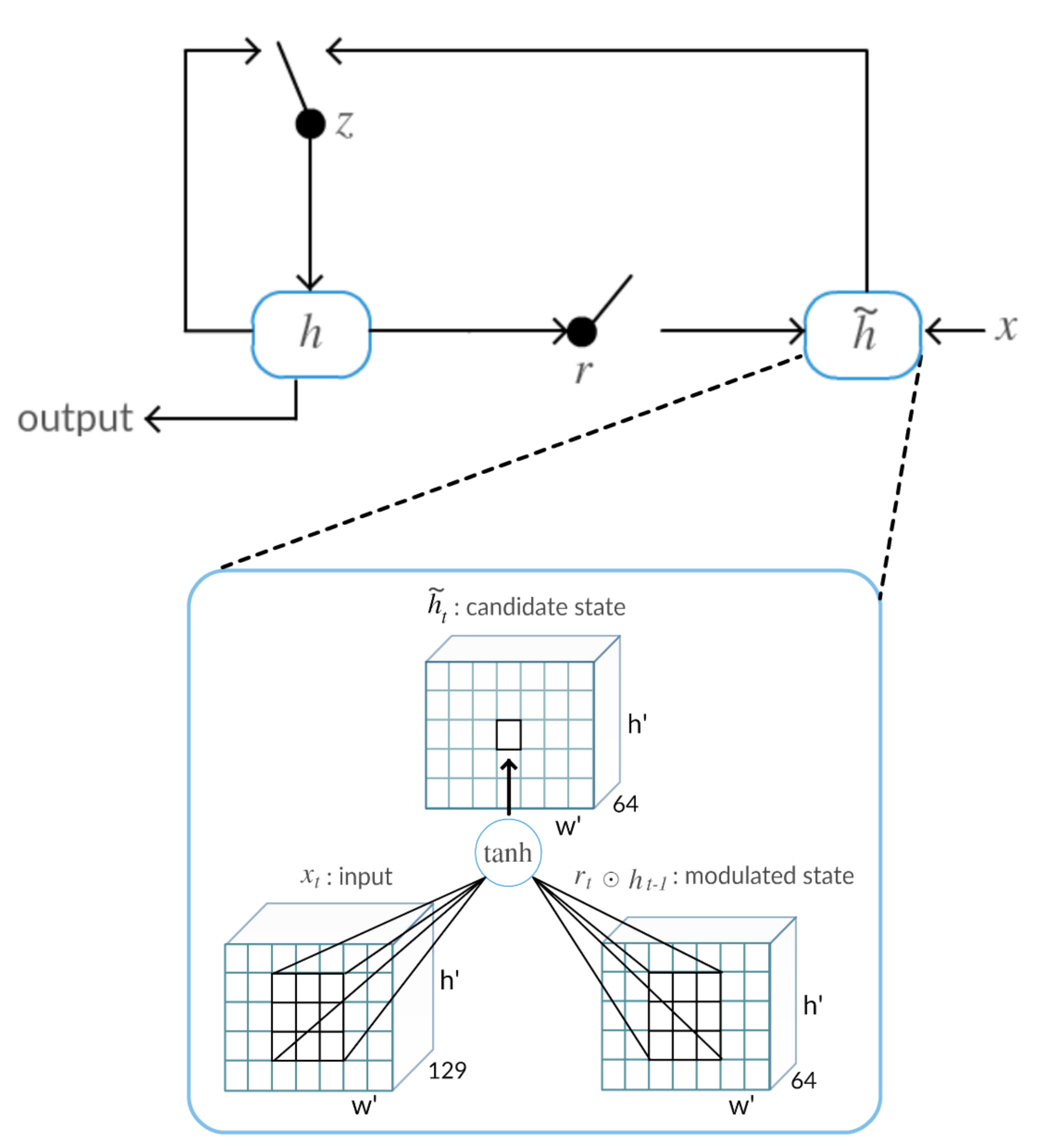}
\end{center}
\vspace{-0.3cm}
\caption{Illustration of ConvGRU with details for the candidate hidden state
module, where  $\tilde{h}_t$ is computed with two convolutional operations and
a $\tanh$ nonlinearity.}
%\vspace{-0.3cm}
\label{fig:lstm}
\end{figure}

The gates and the candidate memory are computed with convolutional operations
over $x_t$ and $h_{t-1}$ shown in equations
(\ref{eqn:update}-\ref{eqn:candmem}). We illustrate the computation of the
candidate memory state $\tilde{h}_t$ in Figure~\ref{fig:lstm}.  The state at
$t-1$, $h_{t-1}$, is first multiplied (element-wise) with the reset gate $r_t$.
This modulated state representation and the input $x_t$ are then convolved with
learned transformations, $w_{h\tilde{h}}$ and $w_{x\tilde{h}}$ respectively,
summed together with a bias term $b_{\tilde{h}}$, and passed though a $\tanh$
nonlinearity. In other words, the visual memory representation of a pixel is
determined not only by the input and the previous state at that pixel, but also
its local neighborhood. Increasing the size of the convolutional kernels allows
the model to handle spatio-temporal patterns with larger motion.

The update and reset gates, $z_t$ and $r_t$, are computed in an analogous
fashion using a sigmoid function instead of $\tanh$. Our ConvGRU applies a
total of six convolutional operations at each time step. All the operations
detailed here are fully differentiable, and thus the parameters of the
convolutions ($w$'s and $b$'s) can be learned in an end-to-end fashion with
back propagation through time~\cite{werbos1990backpropagation}. In summary, the
model learns to combine appearance features of the current frame with the
memorized video representation to refine motion predictions, or even fully
restore them from the previous observations in case a moving object becomes
stationary.

\subsection{Bidirectional processing}
\label{sec:bidirec}
\begin{figure}[t]
\begin{center}
\includegraphics[width=1\columnwidth]{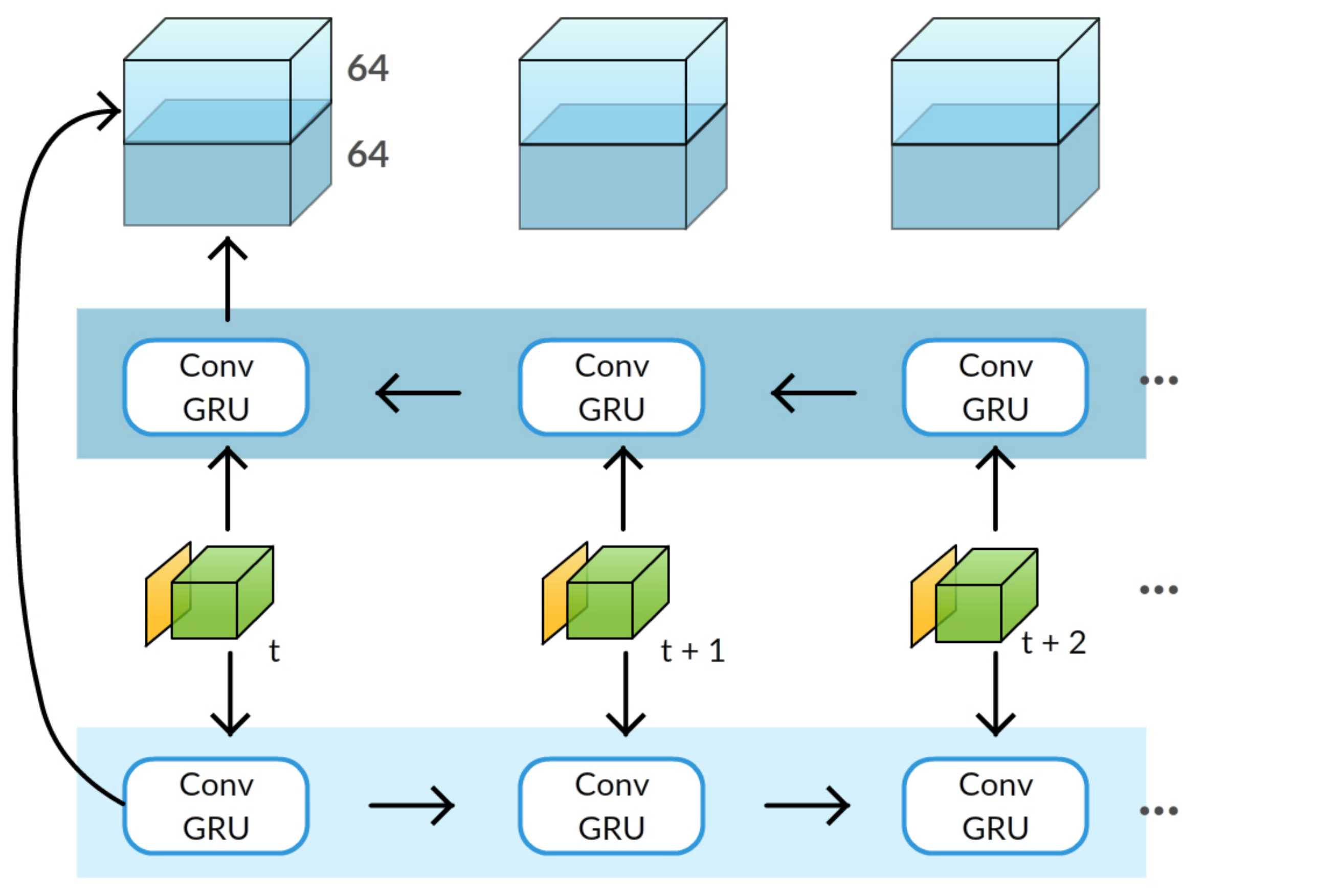}
\end{center}
\vspace{-0.3cm}
\caption{Illustration of the bidirectional processing with our ConvGRU module.}
%\vspace{-0.3cm}
\label{fig:bidir}
\end{figure}
Consider an example where an object is stationary at the beginning of a video
sequence, and starts to move in the latter frames. Our approach described so
far, which processes video frames sequentially (in the forward direction),
starting with the first frame can not segment the object in the initial frames.
This is due to the lack of prior memory representation of the object in the
first frame. We improve our framework with a bidirectional processing step,
inspired by the application of recurrent models bidirectionally in the speech
domain~\cite{graves2005framewise,graves2013hybrid}.

The bidirectional variant of our ConvGRU is illustrated in
Figure~\ref{fig:bidir}. It is composed of two ConvGRU instances with identical
learned weights, which are run in parallel. The first one processes frames in
the forward direction, starting with the first frame (shown at the bottom in
the figure). The second instance process frames in the backward direction,
starting with the last video frame (shown at the top in the figure). The
activations from these two directions are concatenated at each time step, as
shown in the figure, to produce a $128 \times \text{w}/8 \times \text{h}/8$
output. It is then passed through a $3 \times 3$ convolutional layer to finally
produce a $64 \times \text{w}/8 \times \text{h}/8$ output for each frame.
Pixel-wise segmentation from this activation is the result of the last $1
\times 1$ convolutional layer and softmax nonlinearity, as in the
unidirectional case.

Bidirectional ConvGRU is used both in training and in testing, allowing the
model to learn to aggregate information over the entire video. In addition to
handling cases where objects move in the latter frames, it improves the ability
of the model to correct motion prediction errors. As discussed in the
experimental evaluation, bidirectional ConvGRU improves segmentation
performance by nearly 3\% on the DAVIS dataset (see Table~\ref{tbl:abl}). The
influence of bidirectional processing is more prominent on the FBMS dataset, where objects can be static in the beginning of a video,
with 5\% improvement over the unidirectional variant.

\subsection{Training}
\label{sec:train}
We train our visual memory module with the back propagation through time
algorithm~\cite{werbos1990backpropagation}, which unrolls the recurrent network
for $n$ time steps and keeps all the intermediate activations to compute the
gradients. Thus, our ConvGRU model, which has six internal convolutional
layers, trained on a video sequence of length $n$, is equivalent to a $6n$
layer CNN for the unidirectional variant, or $12n$ for the bidirectional model
at training time. This memory requirement makes it infeasible to train the
whole model, including appearance and motion streams, end-to-end. We resort to
using pretrained versions of the appearance and motion networks, and train the
ConvGRU.

In contrast to our motion segmentation model, which is learned on synthetic
videos, we use the training split of the DAVIS dataset~\cite{Perazzi16} for
learning the ConvGRU weights. Despite being an order of magnitude smaller,
DAVIS consists of realistic videos, which turns out to be crucial for effective
use of appearance stream to correct motion estimation errors (see
\S\ref{sec:abl}). Since objects move in all the frames in DAVIS, it biases the
memory module towards the presence of an uninterrupted motion stream. This
results in the ConvGRU learned from this data failing, when an object stops to
move in a test sequence. We augment the training data to simulate such {\it
stop-and-go} scenarios to learn a more robust model for realistic
videos.
To this end, we create additional sequences where we duplicate
  the last frame five times, i.e., we create a part of the video in
  which the object is static. This setting allows ConvGRU to learn how
  to segment objects even if they are static, i.e.,  it explicitly
  memorize the moving object in the initial part of the sequence, and
  then segments it in frames where the motion stops. 
We also augment the training data by duplicating the first five frames
to simulates scenarios where the object is static in the beginning
of a sequence.  

\section{Experiments}

\subsection{Datasets and evaluation}
\label{sec:dataset}
We use five datasets in the experimental analysis: FT3D and DAVIS for training
and test, FusionSeg only for training, and FBMS and SegTrack-v2 only for test.

\paragraph{\bf FlyingThings3D (FT3D).} We train our motion segmentation network
with the synthetic FlyingThings3D dataset~\cite{Mayer16}. It contains videos of
various objects flying along randomized trajectories, in randomly constructed
scenes. The video sequences are generated with complex camera motion, which is
also randomized.  FT3D comprises 2700 videos, each containing 10 stereo frames.
The dataset is split into training and test sets, with 2250 and 450 videos
respectively.  Ground-truth optical flow, disparity, intrinsic and extrinsic
camera parameters, and object instance segmentation masks are provided for all
the videos. No annotation is directly available to distinguish moving objects
from stationary ones, which is required to train our network. We extract this
from the data provided as follows. With the given camera parameters and the
stereo image pair, we first compute the 3D coordinates of all the pixels in a
video frame $t$. Using ground-truth flow between frames $t$ and $t+1$ to find a
pair of corresponding pixels, we retrieve their respective 3D scene points.
Now, if the pixel has not undergone any independent motion between these two
frames, the scene coordinates will be identical (up to small rounding errors).
We have made these labels publicly available on our project
website\footnote{\url{http://thoth.inrialpes.fr/research/mpnet}}. Performance on the test set is measured as the
standard intersection over union score between the predicted segmentation and
the ground-truth masks.

\paragraph{\bf DAVIS.} We use the densely annotated video segmentation
dataset~\cite{Perazzi16} for evaluation as well as for training our visual
memory module. DAVIS contains 50 full HD videos, featuring diverse types of
object and camera motion. It includes challenging examples with occlusion,
motion blur and appearance changes. Accurate pixel-level annotations are
provided for the moving object in all the video frames. A single object is
annotated in each video, even if there are multiple moving objects in the
scene. Following the 30/20 training/validation split provided with the dataset,
we train the visual memory module on the 30 sequences, and test on the 20
validation videos. Note that our motion segmentation model is also evaluated
separately on the entire trainval set, as it is trained exclusively on FT3D.
We evaluate on DAVIS with the three measures used in~\cite{Perazzi16}, namely
intersection over union for region similarity, F-measure for contour accuracy,
and temporal stability for measuring the smoothness of segmentation over time.
We follow the protocol in~\cite{Perazzi16} and use images downsampled by a
factor of two.

\paragraph{\bf FusionSeg.} Jain et al.~\cite{jain2017fusionseg} recently
introduced a dataset containing 84929 pairs of frames extracted from the
ImageNet-Video dataset~\cite{russakovsky2015imagenet}. The frames are annotated
with an automatic segmentation method, which combines a foreground-background
appearance-based model with ground truth bounding box annotations available in
ImageNet-Video. Annotations obtained in this way may be inaccurate, but are
useful for analyzing the impact of learning the motion network on these
realistic examples, in contrast to using synthetic examples; see
Section~\ref{sec:realvidtrain}. We will refer to this dataset as FusionSeg in
the rest of the paper.

\paragraph{\bf FBMS.} The Freiburg-Berkeley motion segmentation
dataset~\cite{ochs2014segmentation} is composed of 59 videos with ground truth
annotations in a subset of frames.  In contrast to DAVIS, it has multiple
moving objects in several videos with instance-level annotations. Also, objects
may move only in a fraction of the frames, but are annotated in frames where
they do not exhibit independent motion. The dataset is split into training and
test set.  Following the standard protocol on this
dataset~\cite{keuper2015motion}, we do not train on any of these sequences, and
evaluate separately on both splits with precision, recall and F-measure
scores. We also convert instance-level annotation to binary ones by merging all
the foreground labels into a single category, as in~\cite{taylor2015causal}.

\paragraph{\bf SegTrack-v2.} It contains 14 videos with instance-level moving
object annotations in all the frames. We convert these annotations into a
binary form for evaluation and use intersection over union as the performance
measure. Note that some videos in this dataset are of very low resolution,
which appears to have a negative effect on the performance of deep learning-based methods trained on high resolution images.

\subsection{Implementation details}

\paragraph{Appearance stream.} For the experiments using DeepLab-v1, we extract
features from the fc6 layer of the network, which has a dilation of 12. This
approach cannot be followed for DeepLab-v2 however, since dilation is applied
to fc8, the prediction layer, in this improved model. Thus, extracting fc6 or
fc7 features of DeepLab-v2 would result in a decreased field of view compared
to the baseline v1 model. Moreover, there are four independent prediction
layers in v2 with dilations 6, 12, 18 and 24, whose outputs are averaged. To
make the feature representation derived from the two architectures compatible,
we introduce four new penultimate convolutional layers to the DeepLab-v2
architecture. These layers have kernel size 3, feature dimension 512  and 
dilations corresponding to those in the prediction layers of DeepLab-v2. The maximum response over these four feature maps is then passed
to a single prediction layer. We finetune this model on PASCAL VOC 2012 for
semantic segmentation. The features after the max operation are used as the
appearance representation in our final model, and correspond to an improved
version of fc6 features from DeepLab-v1. This representation is further passed
through two $1 \times 1$ convolutional layers, interleaved with \textit{tanh}
nonlinearities, to reduce the dimension to 128 for both architectures. 

\paragraph{Training MP-Net.} We use mini-batch SGD with a batch size of 13
images---the maximum possible due to GPU memory constraints. The network is
trained from scratch with learning rate set to $0.003$, momentum to $0.9$, and
weight decay to $0.005$. Training is done for 27 epochs, and the learning rate
and weight decay are decreased by a factor of $0.1$ after every 9 epochs. We
downsample the original frames of the FT3D training set by a factor 2, and
perform data augmentation by random cropping and mirroring. Batch
normalization~\cite{ioffe2015batch} is applied to all the convolutional layers
of the network.

\paragraph{Training visual memory module.} We minimize the binary cross-entropy
loss using back-propagation through time and RMSProp~\cite{rmsprop} with a
learning rate of $10^{-4}$. The learning rate is gradually decreased after
every epoch. Weight decay is set to $0.005$. Initialization of all the
convolutional layers, except for those inside the ConvGRU, is done with the
standard \textit{xavier} method~\cite{glorot2010understanding}. We clip the
gradients to the $[-50, 50]$ range before each parameter update, to avoid
numerical issues~\cite{graves2013generating}. We form batches of size 14 by
randomly selecting a video, and a subset of 14 consecutive frames in it. Random
cropping and flipping of the sequences is also performed for data augmentation.
Our full model uses $7 \times 7$ convolutions in all the ConvGRU operations.
The weights of the two $1 \times 1$ convolutional (dimensionality reduction)
layers in the appearance network and the final $1 \times 1$ convolutional layer
following the memory module are learned jointly with the memory module.  The
model is trained for 30000 iterations and the proportion of batches with
simulated motion discontinuities (see Section~\ref{sec:train}) is set to 20\%.

\paragraph{Other details.} We perform zero-mean normalization of the flow field
vectors, similar to~\cite{simonyan2014two}. When using flow angle and magnitude
together (which we refer to as flow angle field), we scale the magnitude
component, to bring the two channels to the same range.  Our final model uses a
fully-connected CRF~\cite{krahenbuhl2011efficient} to refine boundaries in a
post-processing step. The parameters of this CRF are set to values used for a
related pixel-level segmentation task~\cite{chen2014semantic}. Many sequences
in FBMS are several hundred frames long and do not fit into GPU memory during
evaluation. We apply our method in a sliding window fashion in such cases, with
a window of 130 frames and a step size of 50. Our model is implemented in the
Torch framework.

\subsection{Motion pattern network}
\label{sec:mpnetexp}
We first analyze the different design choices in our MP-Net, and then study
the influence of training data and optical flow representation on the motion
prediction performance.

\subsubsection{Influence of input modalities}
\label{sec:mod}
We analyze the influence of different input modalities, such as RGB data
(single frame and image pair), optical flow field (ground truth and estimated
one) directly as flow vectors, i.e., flow in x and y axes, or as angle field
(flow vector angle concatenated with flow magnitude), and a combination of RGB
data and flow, on training MP-Net. These results are presented on the FT3D test
set and also on DAVIS, to study how well the observations on synthetic videos
transfer to the real-world ones, in Table~\ref{tbl:input}. For computational
reasons we train and test with different modalities on a smaller version of our
MP-Net, with one decoder unit instead of four. Then we pick the best modality
to train and test the full, deeper version of the network.
\begin{table}
\begin{center}
\begin{tabular}{c|l|c|c}
\hline
\# dec. & Trained on FT3D with ... & FT3D & DAVIS \\
\hline
\multirow{6}{*}{1} & RGB single frame  & 68.1 & 12.7  \\
 & RGB pair  & 69.1 & 16.6  \\
 & GT flow  & 74.5 & 44.3  \\
 & GT angle field  & 73.1 & 46.6  \\
 & RGB + GT angle field  & 74.8 & 39.6  \\
 & LDOF angle field  & 63.2 & 38.1  \\
\hline
4 & GT angle field  & 85.9 & 52.4  \\
\hline
\end{tabular}
%\vspace{0.1cm}
\caption{Comparing the influence of different input modalities on the
FlyingThings3D (FT3D) test set and DAVIS. Performance is shown as mean
intersection over union scores. \# dec.\ refers to the number of decoder units
in our MP-Net. Ground-truth flow is used for evaluation on FT3D and LDOF flow
for DAVIS.}
\label{tbl:input}
%\vspace{-0.85cm}
\end{center}
\end{table}

From Table~\ref{tbl:input}, the performance on DAVIS is lower than on FT3D.
This is expected as there is a domain change from synthetic to real data, and
that we use ground truth optical flow as input for FT3D test data, but
estimated flow~\cite{Brox11a,sundaram2010dense} for DAVIS. As a baseline, we
train on single RGB frames (`RGB single frame' in the table). Clearly, no
motion patterns can be learned in this case, but the network performs
reasonably on FT3D test (68.1), as it learns to correlate object appearance
with its motion. This intuition is confirmed by the fact that `RGB single
frame' fails on DAVIS (12.7), where the appearance of objects and background is
significantly different from FT3D.  MP-Net trained on `RGB pair', i.e., RGB
data of two consecutive frames concatenated, performs slightly better on both
FT3D (69.1) and DAVIS (16.6), suggesting that it captures some motion-like
information, but continues to rely on appearance, as it does not transfer well
to DAVIS.

Training on ground-truth flow vectors corresponding to the image pair (`GT
flow') improves the performance on FT3D by 5.4\% and on DAVIS significantly
(27.7\%). This shows that MP-Net learned on flow from synthetic examples can be
transferred to real-world videos. We then experiment with flow angle as part of
the input. As discussed in~\cite{Narayana13}, flow orientations are independent
of depth from the camera, unlike flow vectors, when the camera is undergoing
only translational motion. Using the ground truth flow angle field
(concatenation of flow angles and magnitudes) as input (`GT angle field'), we
note a slight decrease in IoU score on FT3D (1.4\%), where strong camera
rotations are abundant, but in real examples, such motion is usually mild.
Hence, `GT angle field' improves IoU on DAVIS by 2.3\%. We use angle field
representation in all further experiments.

Using a concatenated flow and RGB representation (`RGB + GT angle field')
performs better on FT3D (by 1.7\%), but is poorer by 7\% on DAVIS,
re-confirming our observation that appearance features are not consistent
between the two datasets. Finally, training on computed flow~\cite{Brox11a}
(`LDOF angle field') leads to significant drop on both the datasets: 9.9\% on
FT3D (with GT flow for testing) and 8.5\% on DAVIS, showing the importance of
high-quality training data for learning accurate models. The full version of
our MP-Net, with 4 decoder units, improves the IoU by 12.8\% on FT3D and 5.8\%
on DAVIS over its shallower one-unit equivalent.  

Notice that the performance of our full model on FT3D is excellent, with the
remaining errors mostly due to inherently ambiguous cases like objects moving
close to the camera (see third row in Figure~\ref{fig:qualff}), or very strong
object/camera motion. On DAVIS, the results are considerably lower despite less
challenging motion. To investigate the extent to which this is due to errors
in flow, we study the effect of flow quality in the following section. 
\begin{table}[t]
\begin{center}
\begin{tabular}{l|c|c}
\hline
Flow in test & FT3D & DAVIS \\
\hline
LDOF~\cite{sundaram2010dense}  & 58.7 & 52.4  \\
EpicFlow~\cite{Revaud15} & 52.5 & 56.9  \\
FlowNet 2.0~\cite{ilg2016flownet} & 66.3 & 62.6  \\
\hline
\end{tabular}
\caption{Performance of the best MP-Net variant (4 decoder units trained on GT
angle field) with different flow inputs (LDOF, EpicFlow, FlowNet 2.0) on FT3D
and DAVIS.}
\label{tbl:davis}
\end{center}
\end{table}

\subsubsection{Effect of the flow quality}
\label{sec:realvidexp}
We evaluate the performance of MP-Net using two recent flow estimation methods,
EpicFlow~\cite{Revaud15} and FlowNet 2.0~\cite{ilg2016flownet}, and
LDOF~\cite{Brox11a,sundaram2010dense}, a more classical approach, on the FT3D
test and DAVIS datasets in Table~\ref{tbl:davis}. We observe a significant drop
in performance of 27.2\% (from 85.9\% to 58.7\%) on FT3D when using LDOF,
compared to evaluation with the ground truth in Table~\ref{tbl:input}. This
confirms the impact of optical flow quality and suggests that improvements in
flow estimation can increase the performance of our method on real-world
videos, where no ground truth flow is available.

We experimentally demonstrate this improvement, by utilizing state-of-the-art
flow estimation methods, instead of LDOF. EpicFlow, which leverages motion
contours, produces more accurate object boundaries, and improves over MP-Net
using LDOF by 4.5\% on DAVIS. On FT3D though it leads to a 6.2\% decrease in
performance. We observe that this is due to EpicFlow, which does produce more
accurate object boundaries, but also smooths out small objects and objects with
tiny motions.  This smoothing appears to be beneficial on real videos, but
degrades the performance on synthetic FT3D videos. FlowNet 2.0, which is a
CNN-based method trained on a mixture of synthetic and real videos to estimate
optical flow from a pair of frames, further improves the performance on DAVIS
by 5.7\%. It also achieves better results on FT3D, with a 7.6\% improvement
over LDOF. The remaining gap of 19.6\% between the ground truth flow and
FlowNet 2.0 performance on FT3D shows the potential for future improvement of
flow estimation methods.
\begin{table}[t]
\begin{center}
\begin{tabular}{l|c|c}
\hline
Trained on & FT3D & DAVIS \\
\hline
FT3D & 85.9 & 62.6  \\
FusionSeg & 40.8 & 60.4  \\
FT3D + FusionSeg & 43.0 & 63.9  \\
DAVIS & 34.0 & 62.3  \\
FT3D + DAVIS & 45.7 & 66.7  \\
FT3D + FusionSeg + DAVIS & 40.8 & 68.6  \\
\hline
\end{tabular}
\caption{Performance of the best MP-Net variant trained with different datasets
on FT3D test and DAVIS validation sets. FlowNet 2.0 is used for flow estimation
on DAVIS both in training and in testing in all these experiments.}
\label{tbl:fusionseg}
\end{center}
\end{table}

\subsubsection{Training on real videos}
\label{sec:realvidtrain}
We also experiment with training our MP-Net on FusionSeg and DAVIS, in order to
explore the value real videos can bring in learning a motion segmentation
model, compared to training exclusively on synthetic videos. On one hand, real
videos contain motion patterns that have similar statistics to those
encountered in the testing phase. On the other hand, no ground truth flow
is available, so a noisy flow estimation has to be used, which was shown to be suboptimal when
training on FT3D (see \S\ref{sec:mod}). For FusionSeg the labels are, furthermore, not ground truth, but are instead obtained automatically and contain a significant amount of noise, as discussed in \S\ref{sec:dataset}.

All the models in this experiment are trained on flow extracted with the
state-of-the-art FlowNet 2.0, in order to
minimize the influence of errors in flow. FlowNet 2.0 is also used for
evaluation on the DAVIS validation set, whereas ground truth flow is used for
FT3D test set. As shown in Table~\ref{tbl:fusionseg}, the model trained on
FusionSeg is 2.2\% below the one trained on synthetic data in the case of
DAVIS. On FT3D, its performance drops by 45.1\%. This shows that the synthetic
dataset contains a lot more challenging motions than those typically
encountered in real videos, and although a model learned on synthetic data can
generalize to real data, the converse does not hold. Learning the model only on
real videos also does not bring any improvement on DAVIS, due to errors in flow
estimation and labels in FusionSeg outweighing the potential benefits. We then
finetune the FT3D-trained model on FusionSeg to leverage the benefits of the
two domains. This leads to a notable improvement on both datasets, e.g., 3.5\%
on DAVIS compared to the model trained on FusionSeg alone. The results on
synthetic FT3D videos, despite the improvement over the FusionSeg-trained
model, remain low however, showing the significant difference between the two
domains.

To further explore the use of real videos, we train our motion estimation model
on the DAVIS training set. This dataset contains only 2079 frames, compared to
84929 in FusionSeg, but they are manually annotated, removing one source of
errors due to incorrect labels from training. The performance on DAVIS
increases by 1.9\% with this, compared to training on FusionSeg. On FT3D,
though, IoU decreases by 6.8\%, because the variety of motions in DAVIS is even
smaller than that seen in FusionSeg. Combining the synthetic and real datasets,
i.e., training on FT3D and finetuning on DAVIS, improves the performance on
both FT3D and DAVIS.  Finetuning the FT3D-trained model with FusionSeg and then
DAVIS training data further improves the performance on the DAVIS validation
set, but results in a drop in the case of FT3D, as the model is even
more different from synthetic data.

\subsection{Video object segmentation framework}

\subsubsection{Ablation study}
\label{sec:abl}
\begin{table}[t]
\begin{center}
\begin{tabular}{l|l|c}
\hline
Aspect & Variant & Mean IoU  \\
\hline
  \multicolumn{2}{l|}{Ours (fc6, ConvGRU, Bidir, DAVIS)}  & 70.1    \\
  \hline
  \multirow{6}{*}{App stream} & no  & 43.5  \\
   & RGB  & 58.3  \\
   & 2-layer CNN  & 60.9 \\
   & DeepLab fc7  & 69.8  \\
   & DeepLab conv5  & 67.7  \\
   & DeepLab-v2  & 72.5  \\
  \hline
  App pretrain & ImageNet only  & 64.1  \\
  \hline
  Motion stream & no  & 59.6  \\
  \hline
  \multirow{3}{*}{Memory module} & no & 64.1  \\
  & ConvRNN  & 68.7  \\  
  & ConvLSTM  & 68.9  \\  
    \hline
  Bidir processing & no & 67.2  \\
  \hline
     \multirow{2}{*}{Train data} & FT3D GT Flow & 55.3  \\
   & FT3D LDOF Flow & 59.6   \\
  \hline
\end{tabular}
\caption{Ablation study on the DAVIS validation set showing variants of
appearance and motion streams and memory module. ``Ours'' refers to the model
using fc6 appearance features together with a motion stream, and a
bidirectional ConvGRU trained on DAVIS.}
\label{tbl:abl}
\end{center}
\end{table}

Table~\ref{tbl:abl} demonstrates the influence of different components of our
approach on the DAVIS validation set. We use the model with DeepLab-v1 appearance stream, ConvGRU memory module, bi-directional processing, motion network trained on FT3D+GT-flow and LDOF used for flow estimation on DAVIS as a baseline. We learn all the models on the training set of DAVIS. First, we study the role of the
appearance stream. As a baseline, we remove it completely (``no'' in ``App
stream'' in the table), i.e., the output of the motion stream is the only input
to our visual memory module. In this setting, the memory module lacks
sufficient information to produce accurate segmentations, which results in an
26.6\% drop in performance compared to the method where the appearance stream
with fc6 features is used (``Ours'' in the table). We then provide raw RGB
frames, concatenated with the motion prediction, as input to the ConvGRU. This
simplest form of image representation leads to a 14.8\% improvement, compared
to the motion only model, showing the importance of the appearance features.
The variant where RGB input is passed through two convolutional layers,
interleaved with $\tanh$ nonlinearities, that are trained jointly with the
memory module (``2-layer CNN''), further improves this. This shows the
potential of learning appearance representation as a part of the video
segmentation pipeline. Next, we compare features extracted from the fc7 and
conv5 layers of the DeepLab model to those from fc6 used by default in our
method. Features from fc7 and fc6 show comparable performance, but fc7 ones are
more expensive to compute. Conv5 features perform significantly worse, perhaps
due to a smaller field of view. Finally, we replace the VGG16-based DeepLab
architecture with the ResNet101-based DeepLab-v2, as described in
Section~\ref{sec:fullmodel}. This improves the performance over DeepLab-v1 by
2.4\%, which is consistent with our previous observations that better
representations directly affect the overall performance of the method. We thus
use DeepLab-v2 appearance stream in our final model.

The importance of appearance network pretrained on the semantic segmentation
task is highlighted by the ``ImageNet only'' variant in Table~\ref{tbl:abl},
where the PASCAL VOC pretrained DeepLab segmentation network is replaced with a
network trained on ImageNet classification. Although ImageNet pretraining
provides a rich feature representation, it is less suitable for the video
object segmentation task, which is confirmed by an 6\% drop in performance.
Surprisingly, the variant of our approach that discards the motion information (``no'' in ``Motion stream''), although being 10.5\% below the baseline, still outperforms many of the state-of-the-art methods on DAVIS (see Table~\ref{tbl:soadavis}). This variant learns foreground/background segmentation, which is sufficient for videos with a
single dominant object, but fails in more challenging cases.
Section~\ref{sec:mos} presents additional experiments to explore the quality of
motion estimation during the training and testing phases.

Next, we evaluate the design choices in the visual memory module. We replaced
the memory module (ConvGRU) with a stack of six convolutional layers to obtain
`no memory' variant of our model (``no'' in ``Memory module'' in
Table~\ref{tbl:abl}), but with the same number of parameters. This variant
results in a 6\% drop in performance compared to our full model with ConvGRU on
the DAVIS validation set. The performance of the `no memory' variant is
comparable to 63.3, the performance of ``MP-Net+Obj,'' the approach without any
memory (see Table 2 in~\cite{tokmakov2016learning}). Using a simple recurrent
model (ConvRNN) results in a slight decrease in performance. Such simpler
architectures can be used in case of a memory vs segmentation quality trade
off. The other variant using ConvLSTM is comparable to ConvRNN, possibly due to
the lack of sufficient training data. Performing unidirectional processing
instead of a bidirectional one decreases the performance by nearly 3\% (``no''
in ``Bidir processing'').

Lastly, we train two variants (``FT3D GT Flow'' and ``FT3D LDOF Flow'') on the
synthetic FT3D dataset~\cite{Mayer16} instead of DAVIS. Both of them show a
significantly lower performance than our method trained on DAVIS. This is due
to the appearance of synthetic FT3D videos being very different from the
real-world ones. The variant trained on ground truth flow (GT Flow) is inferior
to that trained on LDOF flow because the motion network (MP-Net) achieves a
high performance on FT3D with ground truth flow, and thus our visual memory
module learns to simply follow the motion stream output.

\subsubsection{Influence of the motion network}
\label{sec:mos}
In Sections~\ref{sec:realvidexp} and~\ref{sec:realvidtrain} we have
demonstrated that the performance of MP-Net can be improved by using more
accurate optical flow estimation methods, and finetuning the network on
FusionSeg and DAVIS. Here we explore the influence of these improvements in
motion estimation on our video object segmentation framework. 
In Table~\ref{tbl:mos} we evaluate the best version of our framework so far
(DeepLab-v2 appearance stream, ConvGRU memory module trained on DAVIS,
Bi-directional processing) with baseline and improved versions of MP-Net. The version denoted as `FT3D + LDOF' in the table corresponds to the segmentation model with baseline MP-Net (trained on FT3D only), and LDOF used
for flow estimation on DAVIS, whereas `FSeg + FNet' corresponds to the model with improved MP-Net (finetuned on FusionSeg) and FlowNet 2.0 used for flow estimation. Note
that the variant of MP-Net finetuned on FusionSeg and then on DAVIS, which showed the best results in the Section~\ref{sec:realvidtrain}, leads to a drop in performance of our video segmentation framework when used in training due to
overfitting on the small number of sequences in DAVIS, thus we do not include it in the comparison. We independently evaluate the effect of replacing the baseline MP-Net with the improved one in training and testing on DAVIS.
\begin{table}[t]
\begin{center}
\begin{tabular}{l|l| c |c}
\hline
Train & Test & Mean IoU & + CRF  \\
\hline
   FT3D + LDOF & FT3D + LDOF & 72.5 & 76.8  \\
   FT3D + LDOF & FSeg + FNet & 72.0 & 75.3  \\
  FSeg + FNet & FSeg + FNet  & 73.3 & 78.2  \\  
  FSeg + FNet & FT3D + LDOF  & 70.2 & 76.2  \\
  \hline
\end{tabular}
\caption{Influence of motion stream variants, used in training and test phases on DAVIS. `FT3D + LDOF' corresponds to the segmentation
model with baseline MP-Net (trained on FT3D only), and LDOF used for flow estimation on DAVIS. `FSeg + FNet' is the variant with improved MP-Net (finetuned on
FusionSeg), and FlowNet 2.0 used for flow estimation.}
\label{tbl:mos}
\end{center}
\end{table}

\begin{table*}[t]
\begin{center}
\begin{tabular}{l | c | rrrrrrrrrr}
\hline
\multicolumn{2}{c|}{Measure} & CVOS~\cite{taylor2015causal} & KEY~\cite{lee2011key} & MSG~\cite{brox2010object} & NLC~\cite{Faktor14} & CUT~\cite{keuper2015motion} & FST~\cite{papazoglou2013fast} & MP-Net-F~\cite{tokmakov2016learning} & FSG~\cite{jain2017fusionseg} & ARP~\cite{kohprimary} & Ours \\
\hline
\multirow{3}{*}{$\mathcal{J}$} & Mean   & 48.2 & 49.8 & 53.3 & 55.1 & 55.2 & 55.8 & 70.0 & 70.7 & 76.2 & 78.2  \\
& Recall  & 54.0 & 59.1 & 61.6  & 55.8 & 57.5 & 64.9 & 85.0 & 83.5 & 91.1 & 89.1  \\
& Decay & 10.5 & 14.1 & 2.4 & 12.6 & 2.3 & 0.0 & ~~1.4 & ~1.5 & 7.0 & 4.1  \\
\hline
\multirow{3}{*}{$\mathcal{F}$} & Mean   & 44.7 & 42.7 & 50.8 & 52.3 & 55.2 & 51.1 & 65.9 & 65.3 & 70.6 & 75.9  \\
& Recall   & 52.6 & 37.5 & 60.0 & 51.9 & 61.0 & 51.6 & 79.2 & 73.8 & 83.5 & 84.7  \\
& Decay   & 11.7 & 10.6 & 5.1 & 11.4 & 3.4 & 2.9 & ~~2.5 & ~~1.8 & 7.9 & 3.5  \\
\hline
$\mathcal{T}$ & Mean   & 24.4 & 25.2 & 29.1 & 41.4 & 26.3 & 34.3 & 56.3 & 32.8 & 39.3 & 20.2  \\
\hline
\end{tabular}
\caption{Comparison to the state-of-the-art methods on DAVIS with intersection
over union ($\mathcal{J}$), F-measure ($\mathcal{F}$), and temporal stability
($\mathcal{T}$).}
\label{tbl:soadavis}
%\vspace{-0.4cm}
\end{center}
\end{table*}

\begin{figure*}[t]
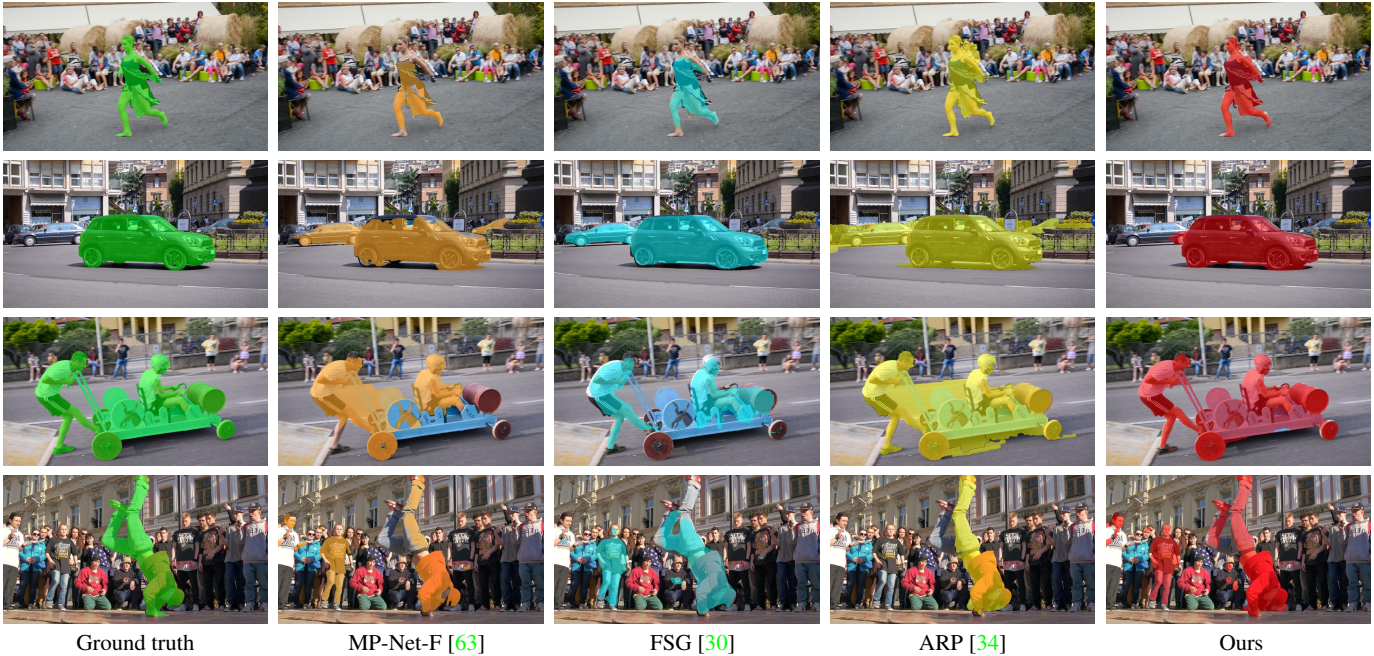

\begin{center}
\fivefigures{soa/dance-twirl/gt}{soa/dance-twirl/mpnet}{soa/dance-twirl/fsg.jpg}{soa/dance-twirl/arp.jpg}{soa/dance-twirl/ours}{0.20\textwidth}\vspace{0.1cm}
\fivefigures{soa/car-roundabout/gt}{soa/car-roundabout/mpnet}{soa/car-roundabout/fsg}{soa/car-roundabout/arp}{soa/car-roundabout/ours}{0.20\textwidth}\vspace{0.1cm}
\fivefigures{soa/soapbox/gt}{soa/soapbox/mpnet}{soa/soapbox/fsg.jpg}{soa/soapbox/arp.jpg}{soa/soapbox/ours}{0.20\textwidth}\vspace{0.1cm}
\fivefigurescaption{soa/breakdance/gt}{soa/breakdance/mpnet}{soa/breakdance/fsg}{soa/breakdance/arp}{soa/breakdance/ours}{0.2\textwidth}
\end{center}
\vspace{-0.3cm}\caption{Qualitative comparison with the top-performing methods
on DAVIS. Left to right: ground truth, results of
MP-Net-F~\cite{tokmakov2016learning}, FSG~\cite{jain2017fusionseg},
ARP~\cite{kohprimary}, and our method.}
\label{fig:davis}
\end{figure*}
The main observation from the results in Table~\ref{tbl:mos} is that our
approach is fairly robust to the motion estimation model being used.
The
performance differs by at most 3\% here, whereas the MP-Net variants
differ by 11.5\%, as seen in Tables~\ref{tbl:davis} and~\ref{tbl:fusionseg}.
This shows that the visual memory module learns to use appearance and temporal
consistency cues to overcome variations in quality of motion estimation.

The performance on the DAVIS validation set is best when the same motion model
is used in the training and the test phases; see the second and the third rows
in Table~\ref{tbl:mos} for a comparison. This is expected because ConvGRU
adapts to the motion model used in training, and suffers from a domain shift
problem, if this model is replaced during the test phase. The variant trained
and tested with the `FSeg + FNet' model (row 3 in the table), which shows the
best performance, with or without the CRF post-processing is used in the
final version of the model.

\subsection{Comparison to the state-of-the-art}
\label{sec:soa}

In this section we compare the best version of our method (DeepLab-v2 appearance stream, ConvGRU memory module trained on DAVIS,
Bi-directional processing, MP-Net finetuned on FusionSeg with FlowNet 2.0 used or flow estimation (FSeg + FNet) and DenseCRF~\cite{krahenbuhl2011efficient} post-processing) to the state-of-the-art methods on 3 benchmark datasets: DAVIS, FBMS and SegTrack-v2.

\paragraph{\bf DAVIS.}
Table~\ref{tbl:soadavis} compares our approach to the state-of-the-art methods
on DAVIS. In addition to comparing our results to the top-performing
unsupervised approaches reported in~\cite{Perazzi16}, we included the results
of recent methods from the benchmark
website:\footnote{\url{http://davischallenge.org/soa_compare.html}}
CUT~\cite{keuper2015motion}, FSG~\cite{jain2017fusionseg} and
ARP~\cite{kohprimary}, as well as the frame-level variant of our method:
MP-Net-F~\cite{tokmakov2016learning}. This frame-level
  approach augments our motion estimation model with an
  heuristic-based objectness score and uses DenseCRF for
  postprocessing (boundary refinement). Our method outperforms
ARP~\cite{kohprimary}, the previous state of the art by 2\% on the mean IoU
measure. We also observe an 8.2\% improvement over MP-Net-F in mean IoU and
36.1\% in temporal stability, which clearly demonstrates the significance of
the visual memory module.

Figure~\ref{fig:davis} shows qualitative results of our approach, and the next
three top-performing methods on DAVIS: MP-Net-F~\cite{tokmakov2016learning},
FSG~\cite{jain2017fusionseg} and ARP~\cite{kohprimary}. In the first row, our
method fully segments the dancer, whereas MP-Net-F and FSG miss various
parts of the person and ARP segments some of the people in the background. All these
approaches use heuristics to combine motion and appearance cues, which become
unreliable in cluttered scenes with many objects. Our approach does not include
any heuristics, which makes it robust to this type of errors. In the second
row, all the methods segment the car, but only our approach does not leak into
other cars in the video, showing high discriminability. In the next row, our
approach is able to fully segment a complex object, whereas the other methods
either miss parts of it (MP-Net-F and FSG) or segment background regions as
moving (ARP). In the last row, we illustrate a failure case of our method. The
people in the background move in some of the frames in this example. MP-Net-F,
FSG and our method segment them to varying extents. ARP focuses on the
foreground object, but misses a part of it.

\paragraph{\bf FBMS.}
As shown in Table~\ref{tbl:bms}, MP-Net-F~\cite{tokmakov2016learning} is
outperformed by most of the methods on this dataset. Our approach based on
visual memory outperforms MP-Net-F by 21.3\% on the test set and by 21.0\%
on the training set according to the F-measure. FST~\cite{papazoglou2013fast}
based post-processing (``MP-Net-V'' in the table) significantly improves the
results of MP-Net-F on FBMS, but it remains below our approach for all
measures. We compare with ARP~\cite{kohprimary} using masks provided by the
authors on the test set. Our method outperforms ARP on this set by
12.2\% on the F-measure. Overall, our method shows a significantly better
performance than all the other approaches in terms of precision, recall and
F-measure. This demonstrates that the visual memory module, in combination with
a strong appearance representation, handles complex video segmentation
scenarios, where objects move only in a fraction of the frames.
\begin{table*}[t]
\begin{center}
\begin{tabular}{c|c|c c c c c c c c c }
\hline
Measure & Set & KEY~\cite{lee2011key} & MP-Net-F~\cite{tokmakov2016learning} & FST~\cite{papazoglou2013fast} & ARP~\cite{kohprimary} & CVOS~\cite{taylor2015causal}  & CUT~\cite{keuper2015motion} & MP-Net-V~\cite{tokmakov2016learning} & Ours \\
\hline
\multirow{2}{*}{$\mathcal{P}$} & Training & 64.9 & 83.0 & 71.3 & - & 79.2 & 86.6 & 69.3 & 89.9 \\
& Test & 62.3 & 84.0 & 76.3 & 76.1 & 83.4 & 83.1 & 81.4 & 93.8 \\
\hline 
\multirow{2}{*}{$\mathcal{R}$} & Training & 52.7 & 54.2 & 70.6 & - & 79.0 & 80.3 & 80.8 & 83.5 \\
 & Test & 56.0 & 49.4 & 63.3 & 66.9 & 67.9  & 71.5 & 73.9 & 75.3 \\
\hline
\multirow{2}{*}{$\mathcal{F}$} & Training & 58.2 & 65.6 & 71.0 & - & 79.3& 83.4 & 74.6 & 86.6 \\
 & Test & 59.0 & 62.2 & 69.2 & 71.3 & 74.9 & 76.8 & 77.5 & 83.5 \\
\hline
\end{tabular}
\caption{Comparison to the state-of-the-art methods on FBMS with precision
($\mathcal{P}$), recall ($\mathcal{R}$), and F-measure ($\mathcal{F}$).}
\label{tbl:bms}
\end{center}
\end{table*}

\begin{figure}[t]
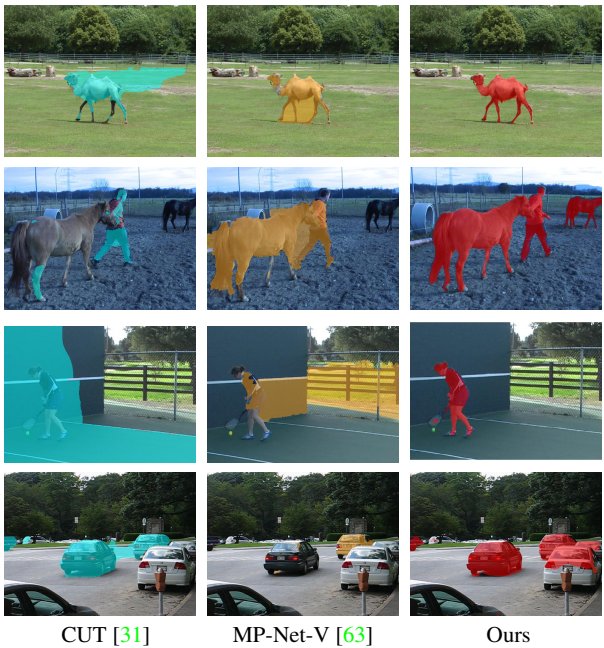

\begin{center}
\threefigures{soafbms/camel/cut}{soafbms/camel/mpnet}{soafbms/camel/ours}{0.3\columnwidth}\vspace{0.1cm}
\threefigures{soafbms/horses04/cut}{soafbms/horses04/mpnet}{soafbms/horses04/ours}{0.3\columnwidth}\vspace{0.1cm}
\threefigures{soafbms/tennis/cut}{soafbms/tennis/mpnet}{soafbms/tennis/ours}{0.3\columnwidth}\vspace{0.1cm}
\threefigurescaption{soafbms/cars2/cut}{soafbms/cars2/mpnet}{soafbms/cars2/ours}{0.3\columnwidth}{CUT~\cite{keuper2015motion}~~}{MP-Net-V~\cite{tokmakov2016learning}}{~~~~~Ours}
\end{center}
\vspace{-0.3cm}\caption{Qualitative comparison with the top-performing methods
on FBMS. Left to right: results of CUT~\cite{keuper2015motion},
MP-Net-Video~\cite{tokmakov2016learning}, and our method.}
\label{fig:fbms}
\end{figure}

Figure~\ref{fig:fbms} shows qualitative results of our method and the two
next-best methods on FBMS: MP-Net-V~\cite{tokmakov2016learning} and
CUT~\cite{keuper2015motion}. MP-Net-V relies highly on
FST's~\cite{papazoglou2013fast} tracking capabilities, and thus leaks
to background in the top three examples, which is a common failure mode of FST.
CUT misses parts of objects and incorrectly assigns background regions to the
foreground in some cases, whereas our method demonstrates very high precision. It is also the only approach which is able to correctly segment all three moving objects in the second example. In the last row we show a failure case of our method. Although it does segment the three moving cars in this video, segmentation leaks to the static cars on the right. Our memory module uses a high-level semantic encoding of the frames to propagate noisy motion segmentations, which leads to incorrectly propagating the segmentation from the moving car to the static ones which are adjacent to it in this case. CUT also captures the three moving cars in this video, but leaks to the background. MP-Net-V does not segment static regions, but misses two of the cars.

\paragraph{\bf SegTrack-v2.}
The performance of our method on SegTrack is presented in the
Table~\ref{tbl:strck}. NLC~\cite{Faktor14} is the top-performing method,
followed by FSG~\cite{jain2017fusionseg}, on this dataset. Note however, that
these methods are both tuned to SegTrack. FSG is trained directly on a subset
of SegTrack sequences, and the parameters of NLC are set manually for this
dataset. In contrast, we use the same model trained on DAVIS in all the experiments, which is
a possible explanation for a lower performance than NLC and FSG. As
shown recently~\cite{jain2017fusionseg,Khoreva16}, the low resolution of some
of the SegTrack videos poses a significant challenge for deep learning-based
video segmentation methods. Being trained on datasets like PASCAL VOC or COCO,
which are composed of high-quality images, these models suffer from the
well-known domain shift problem, when applied to low-resolution videos. Our
method, with its appearance stream trained on VOC, is subject to this issue as
well. Additionally, CRF post-processing decreases the performance of our method
on SegTrack; see `Ours w/o CRF' in Table~\ref{tbl:strck} and qualitative comparison in the next paragraph.
\begin{table}[t]
\begin{center}
\begin{tabular}{l | c}
\hline
Method & Mean IoU \\
\hline
CUT~\cite{keuper2015motion} & 47.8 \\
FST~\cite{papazoglou2013fast} & 54.3 \\
FSG~\cite{jain2017fusionseg} & 61.4 \\
NLC~\cite{Faktor14} & 67.2 \\
Ours & 53.7 \\
Ours w/o CRF & 59.1 \\
\hline
\end{tabular}
\caption{Comparison to the state-of-the-art methods on SegTrack-v2 with mean
IoU.}
\label{tbl:strck}
\end{center}
\end{table}

\begin{figure}[t]
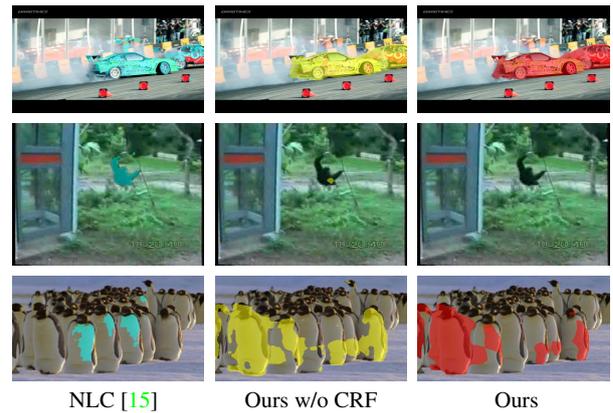

\begin{center}
\threefigures{soasegtrack/drift/nlc}{soasegtrack/drift/nocrf}{soasegtrack/drift/ours}{0.3\columnwidth}\vspace{0.1cm}
\threefigures{soasegtrack/monkeydog/nlc}{soasegtrack/monkeydog/nocrf}{soasegtrack/monkeydog/ours}{0.3\columnwidth}\vspace{0.1cm}
\threefigurescaption{soasegtrack/penguin/nlc}{soasegtrack/penguin/nocrf}{soasegtrack/penguin/ours}{0.3\columnwidth}{NLC~\cite{Faktor14}~~}{Ours w/o CRF}{~~~~~Ours}
\end{center}
\vspace{-0.3cm}\caption{Qualitative comparison of two variants of our method
with the top-performing approach on SegTrack. Left to right: results of
NLC~\cite{Faktor14}, our method without CRF post-processing, and our full
method.}
\label{fig:segtrack}
\end{figure}

A qualitative comparison of our method and the variant without CRF
post-processing (`Ours w/o CRF') with NLC is presented in
Figure~\ref{fig:segtrack}. In the first row, all the three approaches are
segment the moving cars in the challenging racing scene, but NLC is less
precise than the two variants of our method. In the second example, the monkey is fully extracted by NLC only. Our method's prediction (w/o CRF) is
not confident due to the low resolution of the video. It is thus
merged into the background by CRF refinement. In the
last row, none of the methods captures the group of penguins. Our results are
further diminished by the CRF, due to unreliability of the initial prediction
(w/o CRF).

\subsection{ConvGRU visualization}
\label{sec:gru}
\begin{figure*}[t]
\begin{center}
\begin{tabular}{lcc}
 & (a) {\it goat}, $t$ = 23 & (b) {\it dance-twirl}, $t$ = 19 \\
 & \includegraphics[width=0.4\linewidth]{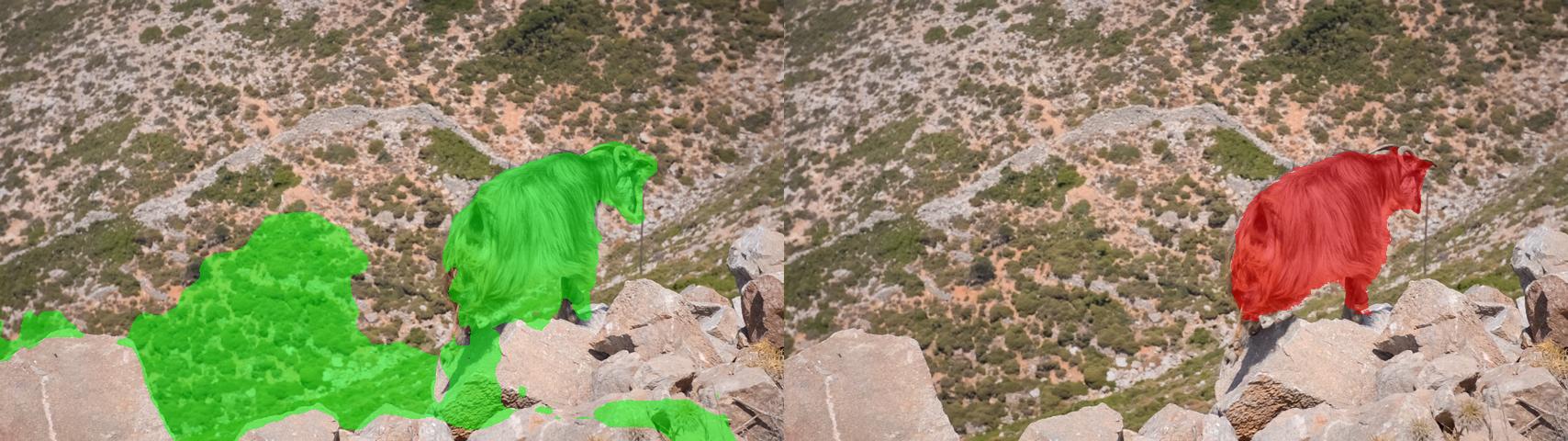} & \includegraphics[width=0.4\linewidth]{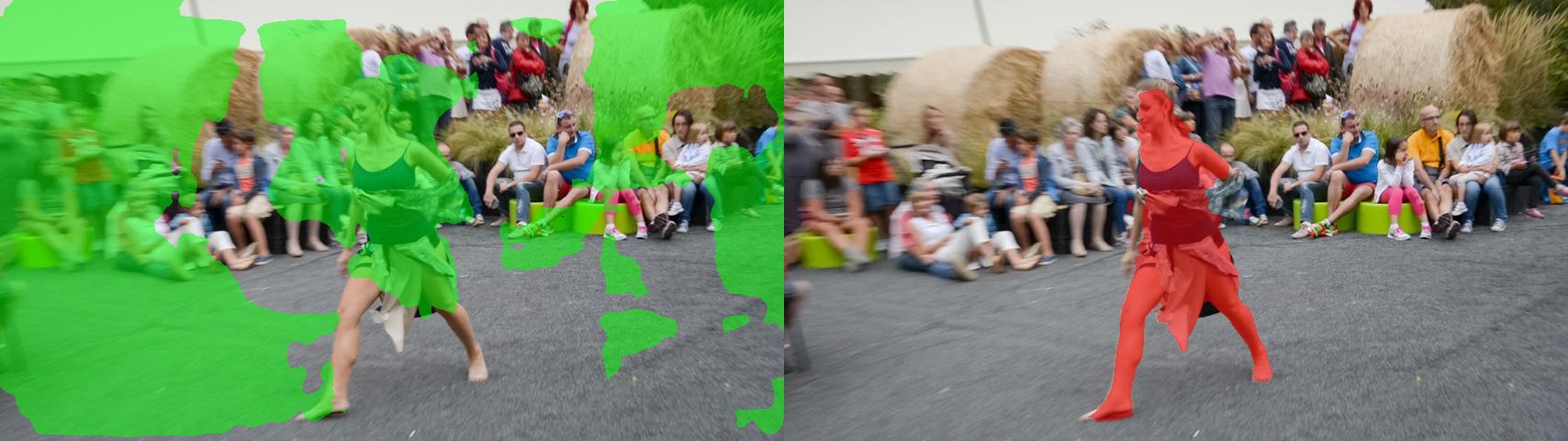}\\
$i =  8$ & \raisebox{-.4\height}{\includegraphics[width=0.4\linewidth]{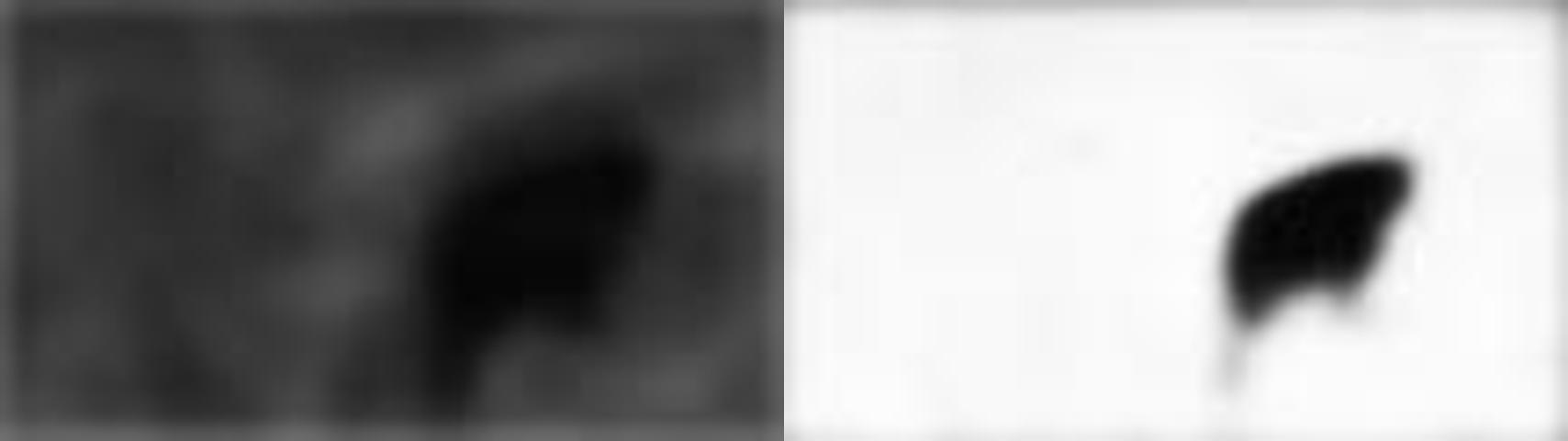}} & \raisebox{-.4\height}{\includegraphics[width=0.4\linewidth]{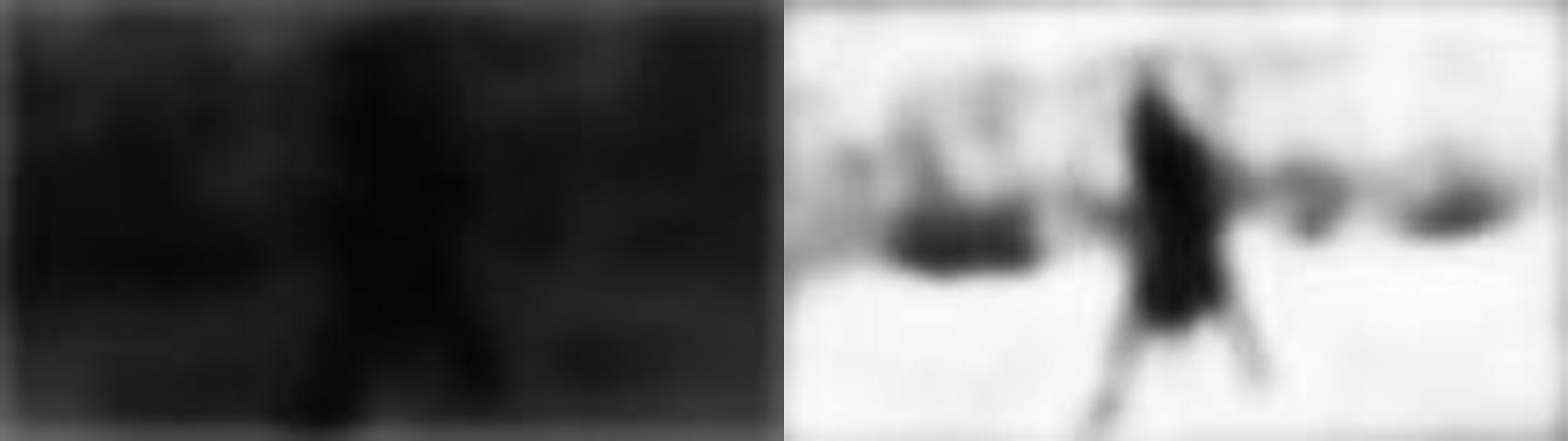}}\vspace{0.1cm}\\
$i = 18$ & \raisebox{-.4\height}{\includegraphics[width=0.4\linewidth]{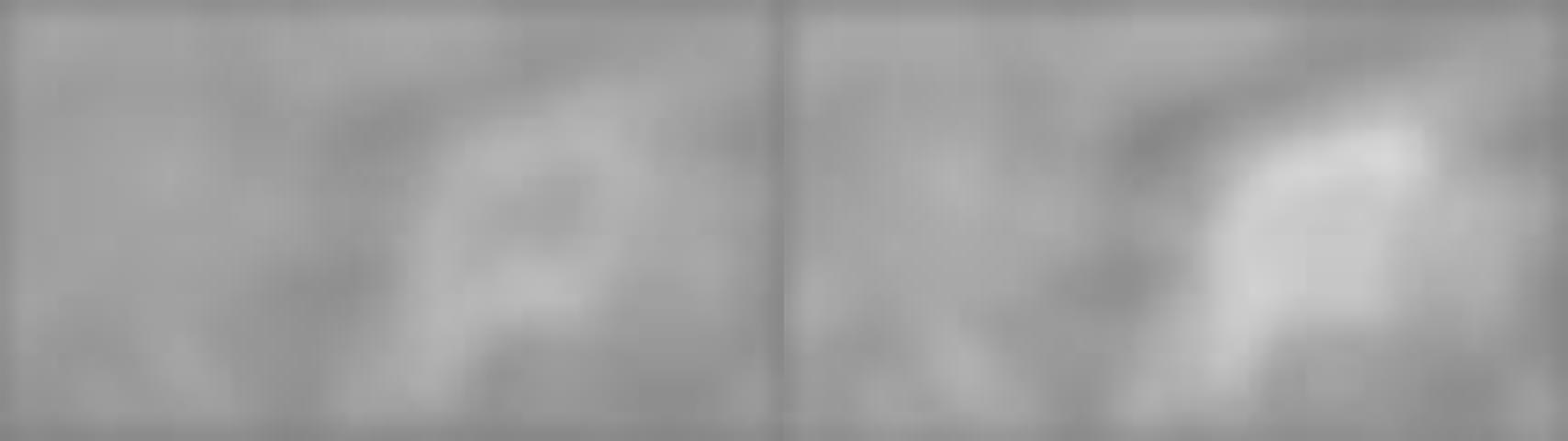}} & \raisebox{-.4\height}{\includegraphics[width=0.4\linewidth]{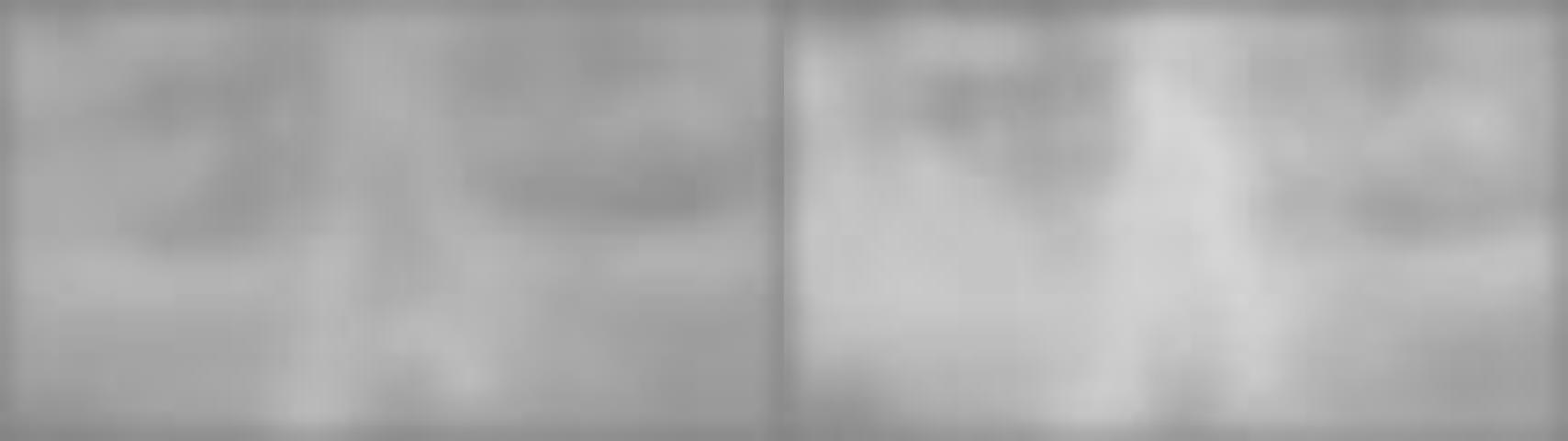}}\vspace{0.1cm}\\
$i = 28$ & \raisebox{-.4\height}{\includegraphics[width=0.4\linewidth]{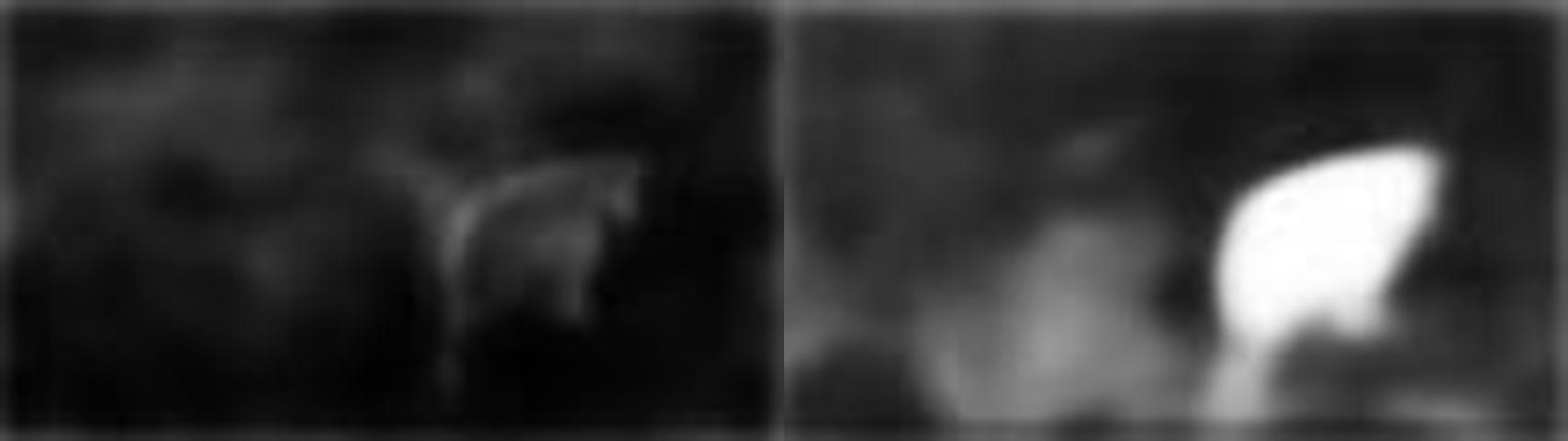}} & \raisebox{-.4\height}{\includegraphics[width=0.4\linewidth]{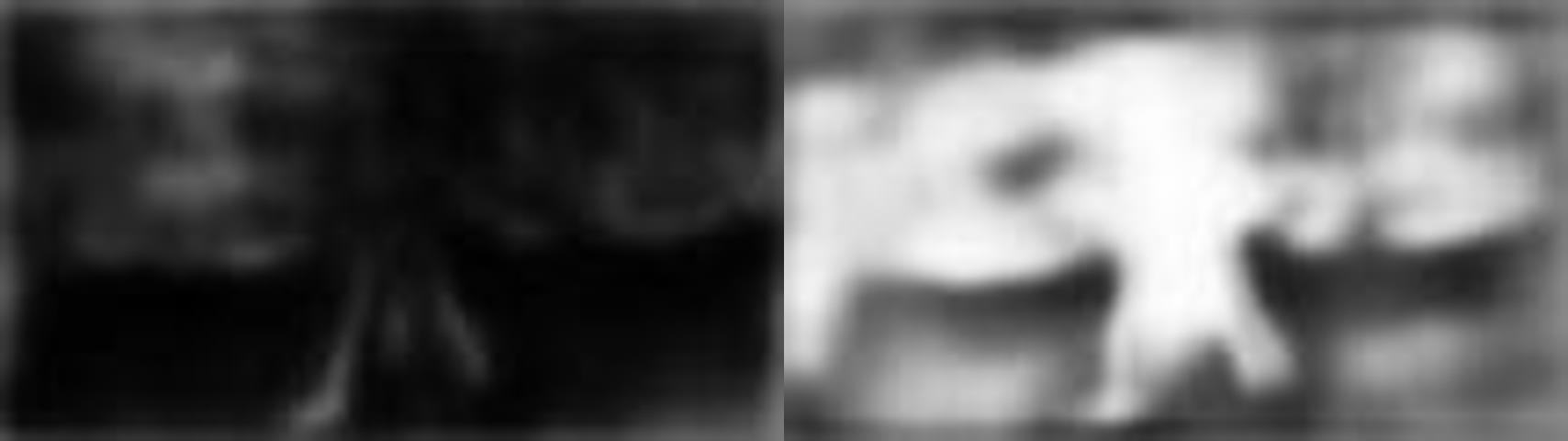}}\vspace{0.1cm}\\
$i = 41$ & \raisebox{-.4\height}{\includegraphics[width=0.4\linewidth]{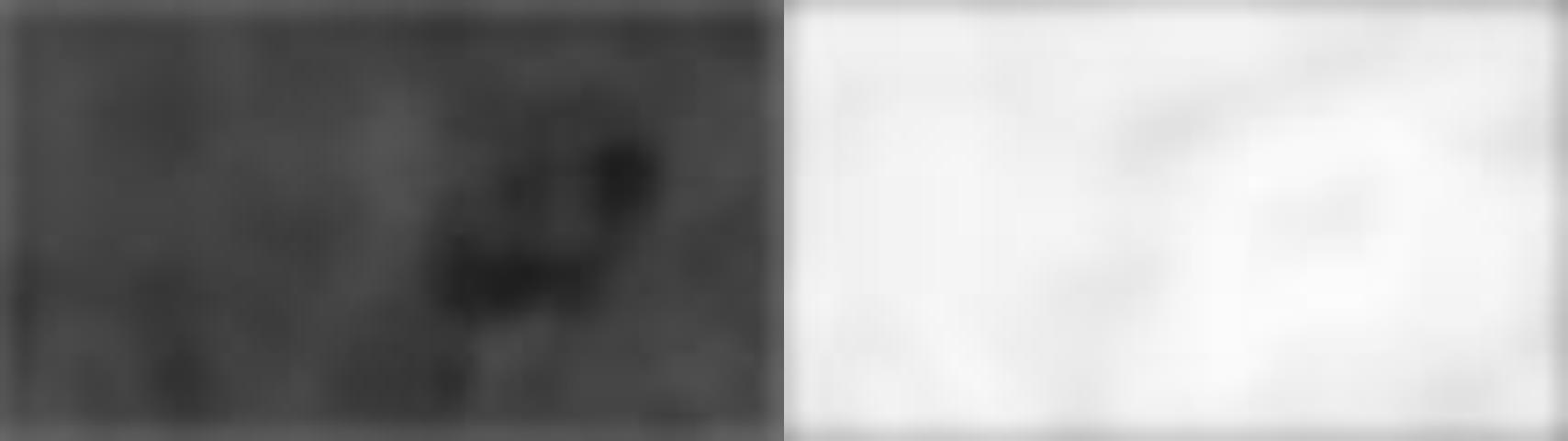}} & \raisebox{-.4\height}{\includegraphics[width=0.4\linewidth]{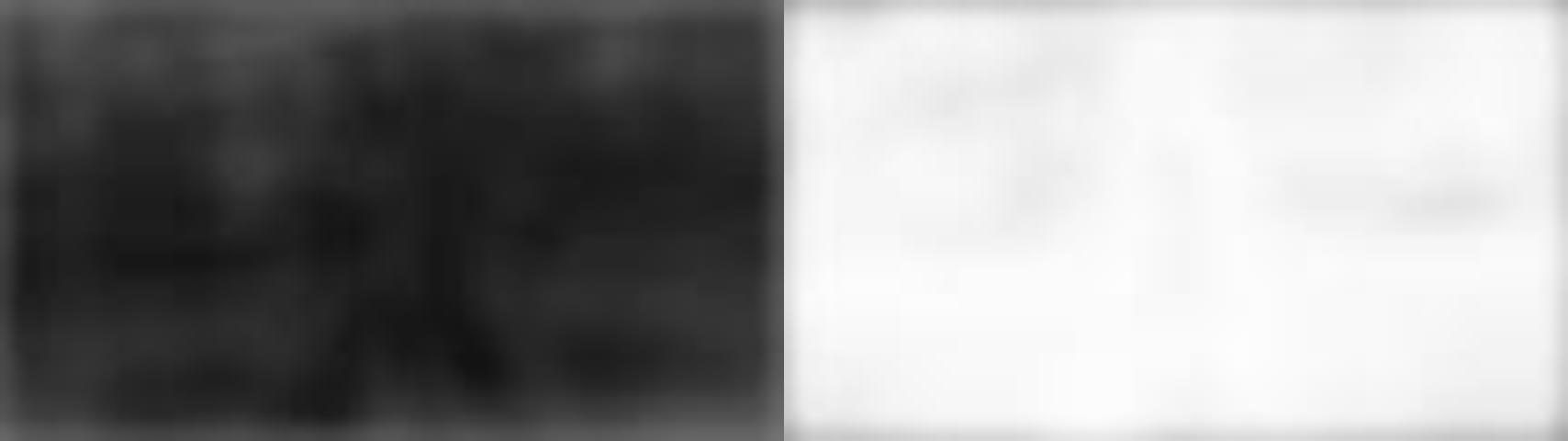}}\vspace{0.1cm}\\
$i = 63$ & \raisebox{-.4\height}{\includegraphics[width=0.4\linewidth]{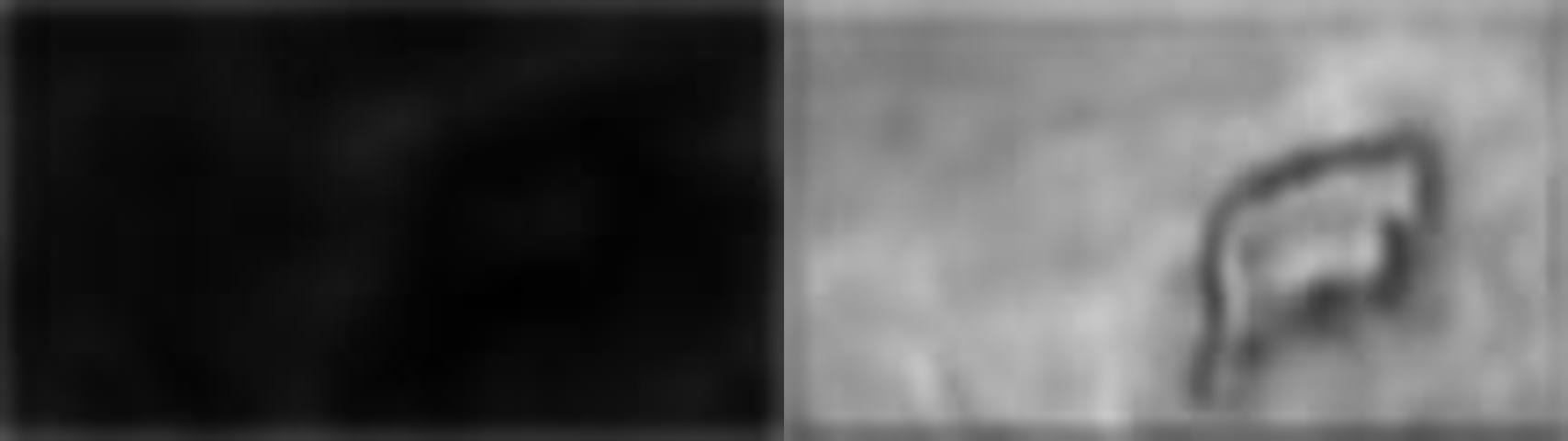}} & \raisebox{-.4\height}{\includegraphics[width=0.4\linewidth]{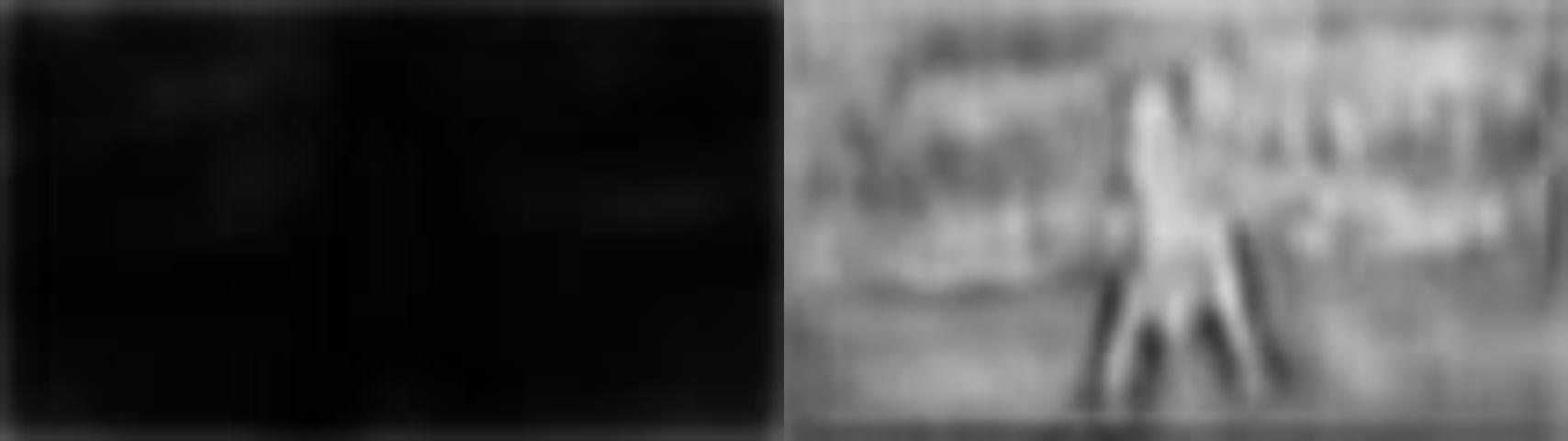}}\vspace{0.1cm}\\
\end{tabular}
\makebox[1\linewidth]{~~~~~~~~~~~~~~~~~~~~~~~~~$r_t^i$ ~~~~~~~~~~~~~~~~~~~~~~~~~~~~~~~~~~~~~$1 - z_t^i$  ~~~~~~~~~~~~~~~~~~~~~~~~~~~~~~~~~~~~~~~~~~~~~$r_t^i$ ~~~~~~~~~~~~~~~~~~~~~~~~~~~~~~~~~~~~~~~$1 - z_t^i$ }
\end{center}
\caption{Visualization of the ConvGRU gate activations for two sequences from
the DAVIS validation set. The first row in each example shows the motion stream output
and the final segmentation result. The other rows are the reset
($r_t$) and the inverse of the update $(1 - z_t)$ gate activations for the corresponding $i$th
dimension. These activations are shown as grayscale heat maps, where white
denotes a high activation.}
\label{fig:gru}
\end{figure*}

We present a visualization of the gate activity in our ConvGRU unit on two
videos from the DAVIS validation set. We use the unidirectional model with the DeepLab-v1 appearance stream and LDOF optical flow in the
following for better clarity. The reset and update gates of the ConvGRU, $r_t$
and $z_t$ respectively, are 3D matrices of $64 \times \text{w}/8 \times
\text{h}/8$ dimension.  The overall behavior of ConvGRU is determined by the
interplay of these 128 components. We use a selection of the components of
$r_t$ and $(1 - z_t)$ to interpret the workings of the gates. Our analysis is
shown on two frames which correspond to the middle of the {\it goat} and {\it
dance-twirl} sequences in (a) and (b), respectively in Figure~\ref{fig:gru}.

The outputs of the motion stream alone (left) and the final segmentation result
(right) of the two examples are shown in the top row in the figure. The five
rows below correspond to one of the 64 dimensions of $r_t$ and $(1 - z_t)$,
with $i$ denoting the dimension. These activations are shown as grayscale heat
maps. High values for either of the two activations increases the influence of
the previous state of a ConvGRU unit on the new state matrix computation. If
both values are low, the state in the corresponding locations is rewritten with
a new value; see equations (\ref{eqn:candmem}) and (\ref{eqn:state}).

For $i=8$, we observe the update gate being selective based on the appearance
information, i.e., it updates the state for foreground objects and duplicates
it for the background. Note that motion does not play a role in this case. This
can be seen in the example of stationary people (in the background) on the
right, that are treated as foreground by the update gate. In the second row,
showing responses for $i=18$, both heatmaps are uniformly close to $0.5$, which
implies that the new features for this dimension are obtained by combining the
previous state and the input at time step $t$.

In the third row for $i=28$, the update gate is driven by motion. It keeps the
state for regions that are predicted as moving, and rewrites it for other
regions in the frame. For the fourth row, where $i=41$, $r_t$ is uniformly
close to 0, whereas $(1 - z_t)$ is close to 1. As a result, the input is
effectively ignored and the previous state is duplicated. In the last row
showing $i=63$, a more complex behavior can be observed, where the gates
rewrite the memory for regions in object boundaries, and use both the previous
state and the current input for other regions in the frame.

\section{Conclusion}
This paper introduces a novel approach for video object segmentation. Our
method combines two complementary sources of information: appearance and
motion, with a visual memory module, realized as a bidirectional convolutional
gated recurrent unit. To separate object motion from camera motion we introduce
a CNN-based model, which is trained using synthetic data to segment
independently moving objects in a flow field. The ConvGRU module encodes
spatio-temporal evolution of objects in a video based on a state-of-the-art
appearance representation, and uses this encoding to improve motion
segmentation. The effectiveness of our approach is validated on three benchmark
datasets. We plan to explore instance-level video object segmentation as part
of future work.

\begin{acknowledgements}
This work was supported in part by the ERC advanced grant ALLEGRO, a Google
research award, a Facebook and an Intel gift. We gratefully acknowledge the
support of NVIDIA with the donation of GPUs used for this work. We also thank
Yeong Jun Koh for providing segmentation masks produced by their
method~\cite{kohprimary} on the FBMS dataset.
\end{acknowledgements}

\bibliographystyle{spmpsci}     
\bibliography{paper}

\end{document}